\documentclass{article} 
\usepackage{iclr2027_conference,times}

\usepackage{amsmath,amsfonts,bm}

\def\eqref#1{equation~\ref{#1}}

\def\1{\bm{1}}

\DeclareMathAlphabet{\mathsfit}{\encodingdefault}{\sfdefault}{m}{sl}
\SetMathAlphabet{\mathsfit}{bold}{\encodingdefault}{\sfdefault}{bx}{n}

\usepackage{hyperref}
\usepackage{url}

\definecolor{catAmb}{RGB}{253,236,220}
\definecolor{catCan}{RGB}{250,224,227}
\definecolor{catGrd}{RGB}{224,231,247}
\definecolor{catTim}{RGB}{224,242,227}
\definecolor{catPha}{RGB}{252,247,219}
\definecolor{catIns}{RGB}{226,243,246}
\definecolor{catSys}{RGB}{237,237,237}
\definecolor{activesaddlerblue}{RGB}{31,119,180}
\newcommand{\ASdiff}[1]{\textcolor{activesaddlerblue}{#1}}

\usepackage{longtable}
\usepackage{booktabs}  
\usepackage{array, makecell}
\usepackage{xcolor}    
\usepackage{pifont}    
\usepackage[utf8]{inputenc} 
\usepackage[T1]{fontenc}    
\usepackage{hyperref}       
\usepackage{url}            
\usepackage{booktabs}       
\usepackage{amsfonts}       
\usepackage{nicefrac}       
\usepackage{microtype}      
\usepackage{xcolor}         
\usepackage{amsmath}
\usepackage{graphicx}
\usepackage{makecell}      
\usepackage[table]{xcolor}  
\definecolor{lightgrayline}{gray}{0.85}  
\usepackage{subcaption}
\usepackage{tcolorbox}
\usepackage{pifont}
\usepackage{multirow}
\usepackage{amssymb}
\usepackage{array} 
\usepackage{enumitem}
\usepackage{makecell}
\usepackage{siunitx}
\tcbuselibrary{breakable, skins}
\usepackage{caption}   
\usepackage{algorithm}
\usepackage{algpseudocode}

\newcommand{\bcirc}[1]{%
  \tikz[baseline=(char.base)]{
    \node[shape=circle, fill=black, inner sep=1.2pt] (char)
      {\color{white}\scriptsize\bfseries #1};
  }%
}

\newif\ifshowcomments
\showcommentstrue 
\ifshowcomments
    \newcommand{\jz}[1]{{\color{blue}[JZ: #1]}}
    \newcommand{\sh}[1]{{\color{purple}[SH: #1]}}
\else
    \newcommand{\jz}[1]{}
    \newcommand{\sh}[1]{}
\fi

\iclrpreprintcopy

\newcommand{\proposed}{ActiveSaddler}

\title{ActiveSaddler: Automated Curriculum Learning for Agent Harness Optimization}

\author{
  Sungho Park\textsuperscript{1}\thanks{Work done during an internship at Microsoft.} \quad
  Wonjoong Kim\textsuperscript{2}\footnotemark[1] \quad
  Jue Zhang\textsuperscript{3}\thanks{Corresponding authors.} \quad
  Wook-Shin Han\textsuperscript{1}\footnotemark[2] \\
  ~\textbf{Pengfei Gao}\textsuperscript{3} \quad
  \textbf{Chanyoung Park}\textsuperscript{2} \quad
  \textbf{Yongqiang Yao}\textsuperscript{3} \quad
  \textbf{Rao Fu}\textsuperscript{3} \\
  ~\textbf{Elsie Nallipogu}\textsuperscript{3} \quad
  \textbf{Qingwei Lin}\textsuperscript{3} \quad
  \textbf{Victor R\"{u}hle}\textsuperscript{3} \\[0.5em]
  \textsuperscript{1}POSTECH \quad
  \textsuperscript{2}KAIST \quad
  \textsuperscript{3}Microsoft \\
  \texttt{juezhang@microsoft.com} \quad
  \texttt{wshan@dblab.postech.ac.kr}
}

\begin{document}

\maketitle

\vspace{-0.5cm}
\begin{abstract}
\vspace{-0.3cm}
Automated harness optimization can substantially improve LLM agents by iteratively updating their prompts, tool interfaces, and control logic from execution feedback. However, existing methods primarily optimize \emph{how} the harness is updated while largely fixing \emph{which training scenarios} generate the feedback that drives those updates. As the harness evolves, the scenarios most useful for further optimization can change, suggesting that the training curriculum itself should adapt alongside the harness. We formulate this missing dimension of harness optimization as an automated curriculum learning problem and introduce \textbf{\proposed}. \proposed{} models the evolving curriculum as a non-stationary bandit with dynamically instantiated optimization targets. It abstracts recurring failures into reusable failure-pattern arms, estimates the potential learning progress from further targeting each pattern, and adaptively balances revisiting known weaknesses with exploring unseen scenarios for new ones. Optimization outcomes continually update both the set of discovered failure patterns and their priorities, allowing the curriculum to co-evolve with the harness. Experiments on GAIA2 and Terminal-Bench 2.0 show that \proposed{} consistently discovers stronger harnesses, improving test Pass@1 by \textbf{4.4} and \textbf{7.5} percentage points over the same harness optimizer using a scenario order fixed before optimization, respectively. Ablations further show that these gains depend on dynamically constructing optimization targets, estimating their evolving utility, and balancing continued optimization with new failure discovery. Together, these results establish automated curriculum learning as a new crucial optimization dimension for harness optimization.
Project website and code will be available at~\href{this URL}{https://aka.ms/ActiveSaddler-website}.

\end{abstract}
\vspace{-0.6cm}
\begin{figure}[!h]
\centering
    \begin{subfigure}[b]{0.49\textwidth}
        \centering
        \includegraphics[width=\linewidth]
        {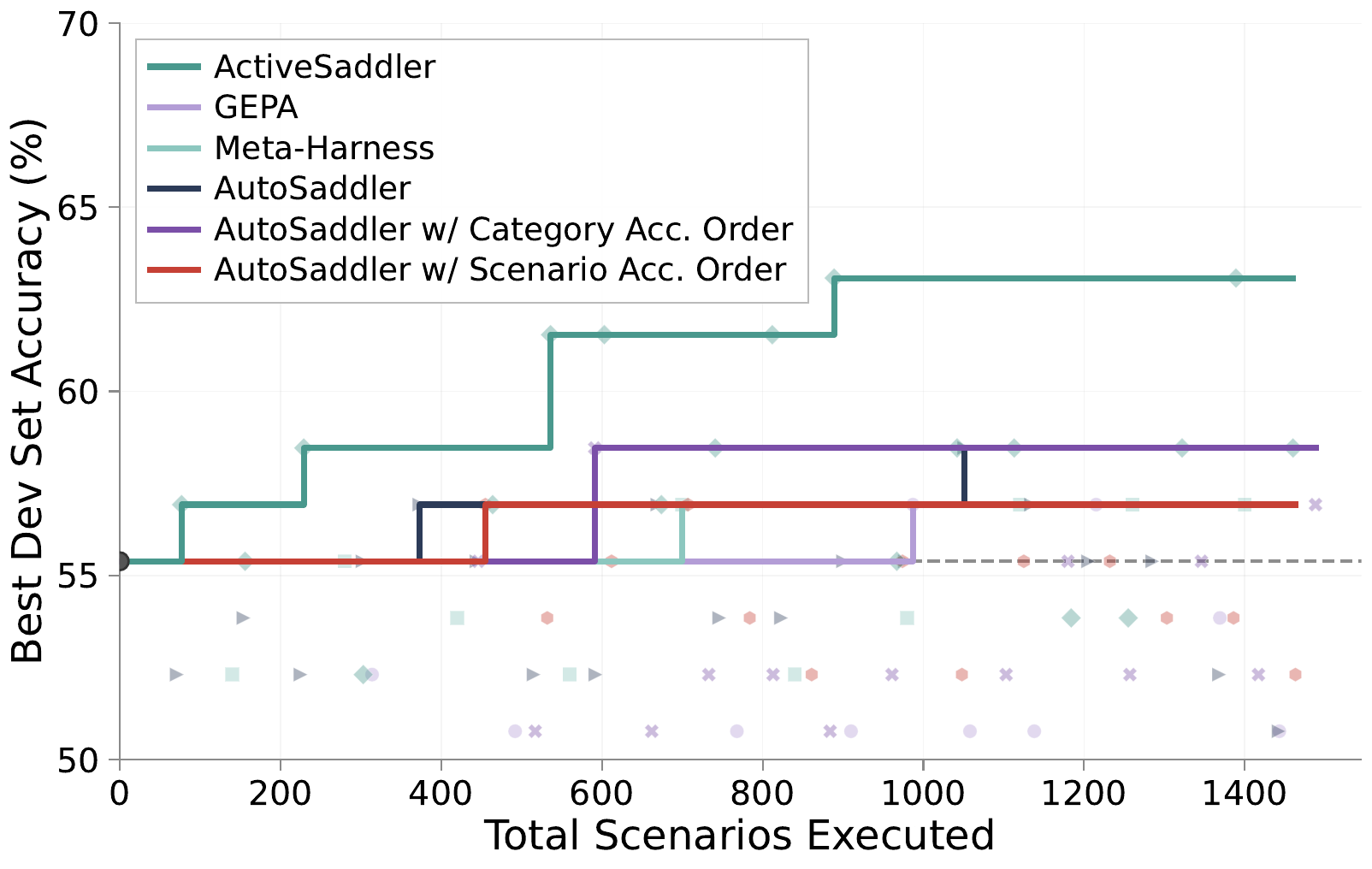}
        \caption{GAIA2.}
        \label{fig:gaia2_curriculum_baseline_comparison}
    \end{subfigure}
    \hfill
    \begin{subfigure}[b]{0.49\textwidth}
        \centering
        \includegraphics[width=\linewidth]
        {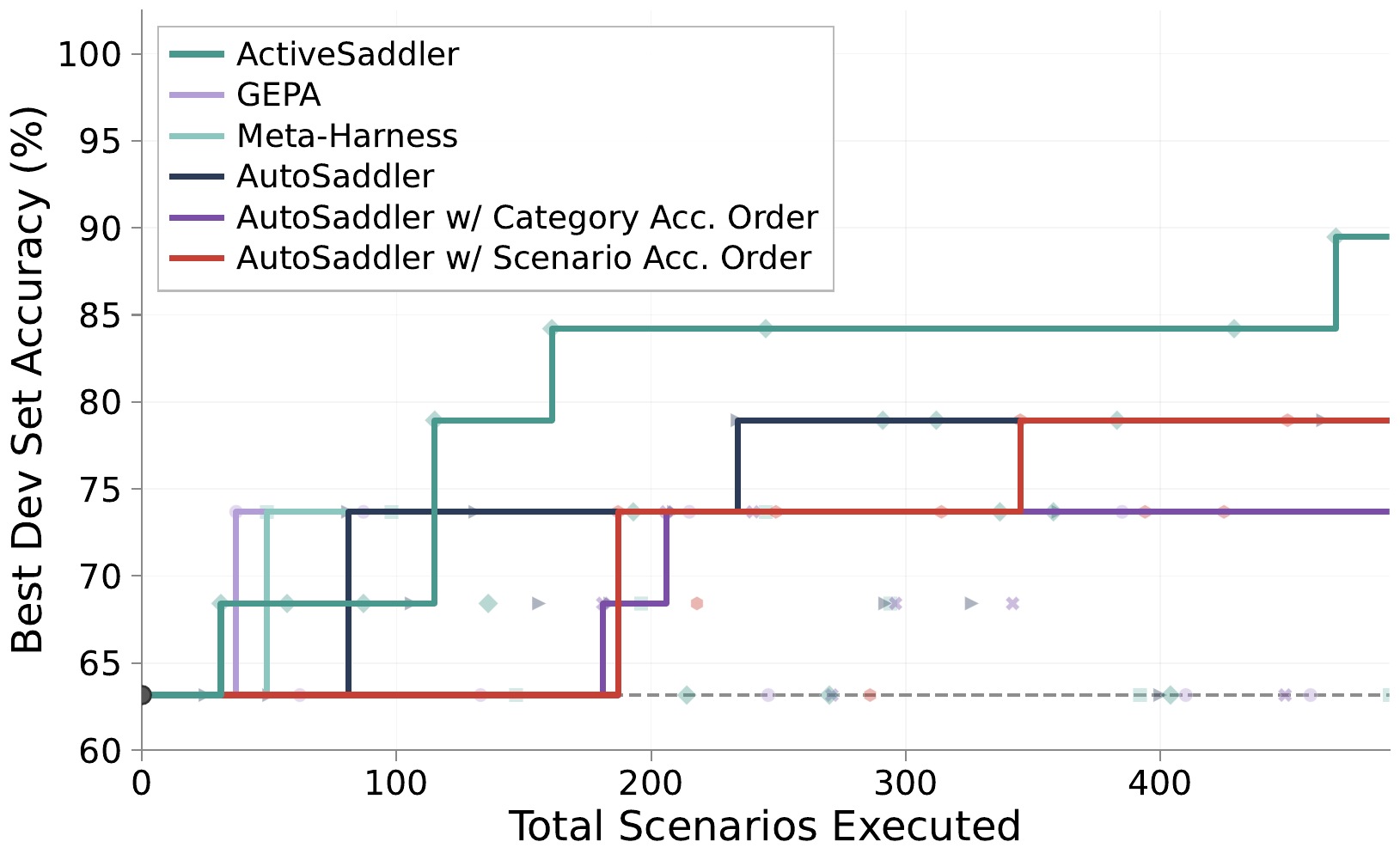}
        \caption{Terminal-Bench 2.0.}
        \label{fig:tbench2_curriculum_baseline_comparison}
    \end{subfigure}
\vspace{-0.2cm}
\caption{
    \textbf{Optimization efficiency across benchmarks.}
    Best-so-far development accuracy versus cumulative task-agent rollouts for
    \proposed{} and baseline methods.
}
\vspace{-0.2cm}
\label{fig:gaia2_curriculum_analysis}
\end{figure}


\section{Introduction}
\label{sec:introduction}

Large Language Models (LLMs) have become increasingly capable of powering autonomous agents, yet reliable performance on multi-step and long-horizon tasks still depends heavily on the external \emph{harness} surrounding the model.
A harness specifies how the agent is prompted, which tools it can use, and how its execution is controlled, monitored, and corrected.
While well-designed harnesses can substantially improve agent robustness~\citep{zhou2026externalization, anthropic-harness-practice-1, openai-harness-practice, langchain-harness-practice}, 
their effectiveness can degrade as models, task distributions, or deployment environments change.
This has motivated automated harness optimization methods that execute training scenarios, diagnose deficiencies from the resulting execution traces, and patch prompts, tool interfaces, or runtime control logic~\citep{lee2026meta, autosaddler}.


These methods thus focus primarily on \emph{how} to update the harness, whereas the training scenarios used to generate feedback are usually predetermined, either as the full training set or as mini-batches scheduled before optimization begins. As the harness evolves, however, the usefulness of these training scenarios can change, creating two limitations. First, unresolved failures may benefit from further optimization, whereas repaired ones may require little additional attention. A fixed schedule cannot respond to this shift and may therefore move on from unresolved failures too early or spend rollouts on scenarios with little remaining value. Indeed, under fixed-order optimization, many failures observed during training remain unresolved by the final optimized harness (\autoref{fig:rq2_quantitative_analysis}).

Second, the relative value of revisiting known failures versus evaluating unseen scenarios also changes over optimization. Revisiting is more valuable when actionable unresolved failures remain, whereas exploring unseen scenarios becomes more valuable when few worthwhile targets remain among the observed failures. A fixed schedule cannot adapt this balance.
These limitations are particularly costly under a limited rollout budget, where optimization opportunities must be allocated selectively. This perspective is consistent with prior work on curriculum and environment design, which shows that learning can benefit from adapting \emph{what} data, tasks, or environments are presented to the learner~\citep{clune2019ai, wang2019paired, DBLP:conf/icml/GravesBMMK17, DBLP:journals/corr/MatiisenOCS17}. This suggests a complementary dimension for harness optimization: as the harness evolves, the curriculum can also adapt \emph{which training scenarios} are used to drive subsequent updates.

In this work, we formulate scenario selection for budgeted offline harness optimization as an automated curriculum-learning problem and introduce \textbf{\proposed}, which \textit{proactively} adapts the optimization curriculum as the harness evolves.
We model curriculum selection as a non-stationary bandit whose arms are instantiated online from observed failures: each \emph{failure-pattern arm} represents a potentially repairable harness deficiency shared across failed trajectories.
At each iteration, \proposed{} either evaluates unseen scenarios to discover new failure patterns or revisits an existing arm, prioritizing arms by their estimated potential for further improvement.
The resulting optimization outcomes update both the priorities of existing arms and the set of discovered patterns.
Through this feedback loop, \proposed{} co-evolves the curriculum with the harness to allocate a fixed rollout budget toward more valuable optimization opportunities.

We evaluate \proposed{} on GAIA2~\citep{GAIA2} and Terminal-Bench 2.0~\citep{TerminalBench2} against existing harness optimizers and fixed curriculum baselines.
Across both benchmarks, \proposed{} achieves the strongest final harness performance, improving over the same harness optimizer with a scenario order fixed before optimization by \textbf{4.4} and \textbf{7.5} percentage points, respectively.
Our ablation analyses reveal three design principles for automated curriculum learning in harness optimization: i) \textbf{failure-pattern arms} provide an effective optimization abstraction: category-level arms can conflate distinct failures, whereas scenario-level arms fragment recurring failures across individual scenarios; ii) \textbf{adaptive arm prioritization} is crucial because the value of revisiting a failure changes as the harness evolves; iii) \textbf{adaptive exploration} is necessary because the relative value of revisiting observed failures versus evaluating unseen scenarios changes over optimization.

Our main contributions are summarized as follows:
\begin{itemize}[leftmargin=1em,itemsep=0.25em,topsep=0.25em]
    \item We formulate scenario selection for offline harness optimization as an automated curriculum-learning problem, where training scenarios are selected adaptively as the harness evolves.
    \item We introduce \proposed{}, which combines failure-pattern arms, adaptive arm prioritization, and adaptive exploration to construct this evolving curriculum.
    \item We show consistent gains on GAIA2 and Terminal-Bench 2.0 over existing harness optimizers and fixed curricula under the same rollout budget.
\end{itemize}

\section{Related Work}
\label{sec:relatedwork}

\paragraph{Automatic Harness Optimization.}
Recent work extends automatic optimization beyond prompts to the broader agent \emph{harness}, including tools, memory, skills, and runtime control logic~\citep{zhou2026externalization}.
Despite implementation differences, these methods generally execute the current harness, use rollout feedback to propose updates, and evaluate candidate harnesses.
Broadly, automatic harness optimization (AHO) spans search-based and learning-based approaches.
Search-based methods explore alternative harness implementations or configurations through evolutionary or Bayesian search~\citep{lee2026meta,lin2026agentic,zhang2026harnesscompass,sengupta2026harbor}.
Learning-based methods instead treat external harness state as the object being learned, repeatedly updating it from rollout feedback on training mini-batches, analogous to model training but without updating model weights~\citep{autosaddler,DBLP:journals/corr/abs-2605-23904}.
GEPA~\citep{agrawal2026gepa} lies between these styles, coupling reflective updates from training mini-batches with evolutionary Pareto search over validation performance.
Recent work further extends AHO by changing the optimization signal, such as recalibrating historical experience or replacing labeled validation with self-preference~\citep{guo2026drevo,pan2026evolving}, and by jointly optimizing harnesses and model weights~\citep{hebbar2026sia,chen2026co,kim2026whale}.
Orthogonally, offline approaches optimize the harness before deployment, often using training data and held-out validation~\citep{autosaddler,DBLP:journals/corr/abs-2605-23904}, whereas online or test-time approaches continue adapting during sequential interaction or evaluation~\citep{DBLP:journals/corr/abs-2605-09998,DBLP:journals/corr/abs-2608-15071,DBLP:journals/corr/abs-2607-08124}.
We focus on offline, learning-based AHO under a standard train/dev/test protocol, where \proposed{} leaves the harness-update mechanism unchanged and instead adapts \emph{which training scenarios} provide evidence for each update.


\paragraph{Automated Curriculum Learning.}
Curriculum learning studies how training data, tasks, or environments are scheduled to improve learning~\citep{DBLP:conf/icml/BengioLCW09}.
Early work prioritized learning opportunities according to competence or learning progress~\citep{DBLP:journals/tec/OudeyerKH07,DBLP:conf/icdl-epirob/LopesO12}, while subsequent automated curricula formalized this choice through non-stationary bandits, teacher-student task selection, and progress-driven environment sampling~\citep{DBLP:conf/icml/GravesBMMK17,DBLP:journals/corr/MatiisenOCS17,DBLP:conf/corl/PortelasCHO19}.
These methods estimate task utility from observable learner-side signals, such as loss reduction, competence progress, or TD error~\citep{DBLP:conf/icml/GravesBMMK17,DBLP:journals/corr/MatiisenOCS17,DBLP:conf/icml/JiangGR21}.
\proposed{} adopts the same principle that training opportunities should be prioritized according to their evolving value to the learner.
Harness optimization, however, makes both the curriculum target and its progress implicit.
A training scenario does not specify a fixed harness weakness: the weakness becomes apparent only after execution, the same scenario may expose different failures as the harness changes, and the same weakness may recur across different scenarios.
Moreover, a diagnosed weakness has no direct loss or other target-level learning signal indicating whether it remains unresolved or worth another intervention; this must instead be inferred from the current harness, prior optimization attempts, and execution history.
\proposed{} therefore constructs failure-pattern arms from diagnosed failures, continually re-estimates their optimization value, and allocates rollouts between revisiting known weaknesses and exploring unseen scenarios for new ones.

\paragraph{Adaptive Task Selection for Agent Optimization.}
Recent work has begun to adapt task allocation at different stages of agent optimization.
For model-weight optimization, CoEvolve~\citep{DBLP:conf/acl/YangMHHWC26} uses forgetting and uncertainty observed in agent rollouts to generate new training tasks targeting behaviors the current agent still struggles with.
In harness optimization, Task-CoEvolve~\citep{DBLP:journals/corr/abs-2608-20169} adaptively selects validation tasks that best distinguish candidate harnesses, while HarnessLens~\citep{DBLP:journals/corr/abs-2608-27311} selects verification tasks relevant to each proposed harness modification.
These methods adapt task selection on the evaluation side of harness optimization, after candidate modifications have been proposed.
In contrast, \proposed{} adapts the training scenarios that generate evidence for harness updates throughout optimization, repeatedly deciding which known weaknesses to revisit and when to execute unseen scenarios to discover new ones.



\section{Preliminaries}
\label{sec:preliminaries}

\paragraph{Harness Optimization.}
Let $H_\theta$ denote an agent harness parameterized by $\theta\in\Theta$, where $\theta$ may include prompts, tool interfaces, and runtime control logic.
For a task scenario $(x,y^\ast)$ drawn from a task distribution $\mathcal{T}$, where $x$ denotes the task input and $y^\ast$ its reference answer, executing the agent with $H_\theta$ produces an execution trace $\tau$ and a final answer $\hat{y}$ according to
$
    (\tau,\hat{y})
    \sim
    P_\theta(\cdot\mid x),
$
where $P_\theta$ denotes the stochastic execution distribution induced by the model and harness. Let $\mu(\hat{y},y^\ast)$ be a task-level evaluation metric. The population performance of a harness is
\begin{equation}
    J(\theta)
    =
    \mathbb{E}_{(x,y^\ast)\sim\mathcal{T}}
    \mathbb{E}_{(\tau,\hat{y})\sim P_\theta(\cdot\mid x)}
    \bigl[\mu(\hat{y},y^\ast)\bigr].
    \label{eq:population_objective}
\end{equation}
For a finite dataset $D$, we define the empirical harness performance as
\begin{equation}
    \widehat{J}_{D}(\theta)
    =
    \frac{1}{|D|}
    \sum_{(x_i,y_i^\ast)\in D}
    \mu(\hat{y}_i,y_i^\ast),
    \qquad
    (\tau_i,\hat{y}_i)
    \sim P_\theta(\cdot\mid x_i).
    \label{eq:empirical_objective}
\end{equation}

Following the learning-based offline harness-optimization setting of AutoSaddler~\citep{autosaddler}, we assume fixed and mutually disjoint training, development, and test sets, $D_{\mathrm{train}}$, $D_{\mathrm{dev}}$, and $D_{\mathrm{test}}$.
At iteration $t$, let $\theta_t$ denote the current harness parameters and write $H_t:=H_{\theta_t}$.
Given a selected set of training scenarios $X_t\subseteq D_{\mathrm{train}}$, executing $H_t$ produces a collection of execution records
\begin{equation}
    \mathcal{E}_t
    =
    \left\{
        (x,y^\ast,\tau,\hat{y})
        \;\middle|\;
        (x,y^\ast)\in X_t,\;
        (\tau,\hat{y})\sim P_{\theta_t}(\cdot\mid x)
    \right\}.
    \label{eq:execution_records}
\end{equation}
A harness optimizer $\mathcal{O}$ uses these records to diagnose deficiencies in the current harness, construct and evaluate candidate updates, and return the harness retained for the next iteration:
\begin{equation}
    \theta_{t+1}
    \sim
    \mathcal{O}\bigl(\theta_t,\mathcal{E}_t,D_{\mathrm{dev}}\bigr),
    \label{eq:harness_optimizer}
\end{equation}
where $D_{\mathrm{dev}}$ supports candidate selection, regression checks, and generalization checks during optimization, while $D_{\mathrm{test}}$ is reserved for final evaluation. Thus, $X_t$ determines \emph{which evidence} is collected, while $\mathcal{O}$ determines \emph{how it is used} to update the harness.

\paragraph{Automated Curriculum Learning for Harness Optimization.}
Automated curriculum learning adapts the selection of training data to the evolving state of a learner~\citep{DBLP:conf/icml/GravesBMMK17}.
In harness optimization, we regard the current harness as the evolving learner and formulate the selection of $X_t$ as a curriculum-learning problem.
Let $\mathcal{I}_{<t}$ denote the information available before the curriculum decision at iteration $t$, including the current harness and the execution and update history accumulated over previous iterations.
A curriculum policy $\pi$ selects the next training-scenario set according to
\begin{equation}
    X_t
    \sim
    \pi(\cdot\mid\mathcal{I}_{<t}),
    \qquad
    X_t\subseteq D_{\mathrm{train}}.
    \label{eq:curriculum_scenario_selection}
\end{equation}
The selected scenarios are executed under $H_t$, and the resulting harness-optimization step provides feedback for subsequent curriculum decisions.
We hold the optimizer $\mathcal{O}$ and its candidate-evaluation procedure fixed and study how the curriculum policy adapts as the harness evolves.

Let $\hat{\theta}_\pi$ denote the final harness selected by development-set performance from the optimization trajectory induced by policy $\pi$.
Under a fixed rollout budget shared across curriculum policies, the automated curriculum-learning objective is
\begin{equation}
    \pi^\ast
    \in
    \arg\max_{\pi}
    \mathbb{E}\!\left[J(\hat{\theta}_\pi)\right].
    \label{eq:active_harness_objective}
\end{equation}
After optimization, we report the performance of the selected harness on the test set as $\widehat{J}_{D_{\mathrm{test}}}(\hat{\theta}_\pi)$.

\section{Proposed Method: \proposed}
\label{sec:method}

We now present \proposed{}, an iterative curriculum-learning framework for harness optimization. We first introduce the problem formulation and overall workflow, then describe key modules in detail.

\paragraph{Problem Formulation and Design Rationale.}
Harness optimization operates under a limited rollout budget, requiring the curriculum to decide at each iteration which training target should receive the next optimization opportunity. This naturally leads to a sequential resource-allocation problem, which we formulate as a \textit{multi-armed bandit}: each arm represents an optimization target, and selecting an arm directs the next round of optimization toward that target \citep{DBLP:conf/icml/GravesBMMK17, DBLP:journals/corr/MatiisenOCS17}.

To instantiate these optimization targets, we define each arm around a \emph{harness weakness} rather than an individual failed execution, since a patch typically addresses an underlying deficiency that may recur across multiple scenarios and traces. Because such weaknesses are not directly observable, we infer them through failure diagnosis and group failures attributed to the same deficiency into a common \emph{failure-pattern arm}. This lets related failures contribute evidence toward a shared optimization target.

This formulation introduces two practical challenges. First, the arm set is not known in advance: as the harness evolves and executions vary stochastically, new failure patterns may emerge. We therefore construct the arm set during optimization and expand it as new failure patterns are diagnosed. Second, each iteration typically executes only a subset of the training set~\citep{autosaddler, agrawal2026gepa}, leaving some scenarios and their potential failure patterns unobserved. The curriculum must therefore balance \emph{exploitation}, by prioritizing among known failure-pattern arms, with \emph{exploration}, by executing unobserved scenarios that may reveal new weaknesses.

These challenges motivate the three core components of \proposed{}: a \textbf{Failure-Pattern Extractor}, which identifies recurring harness weaknesses and organizes them into failure-pattern arms; an \textbf{Arm Prioritizer}, which selects among known arms for further optimization; and an \textbf{Exploration Controller}, which decides when to explore unseen scenarios that may reveal new weaknesses. We next describe how these components interact within the iterative harness-optimization loop.

\begin{figure}[!t]
    \centering
    \includegraphics[width=0.9\linewidth]{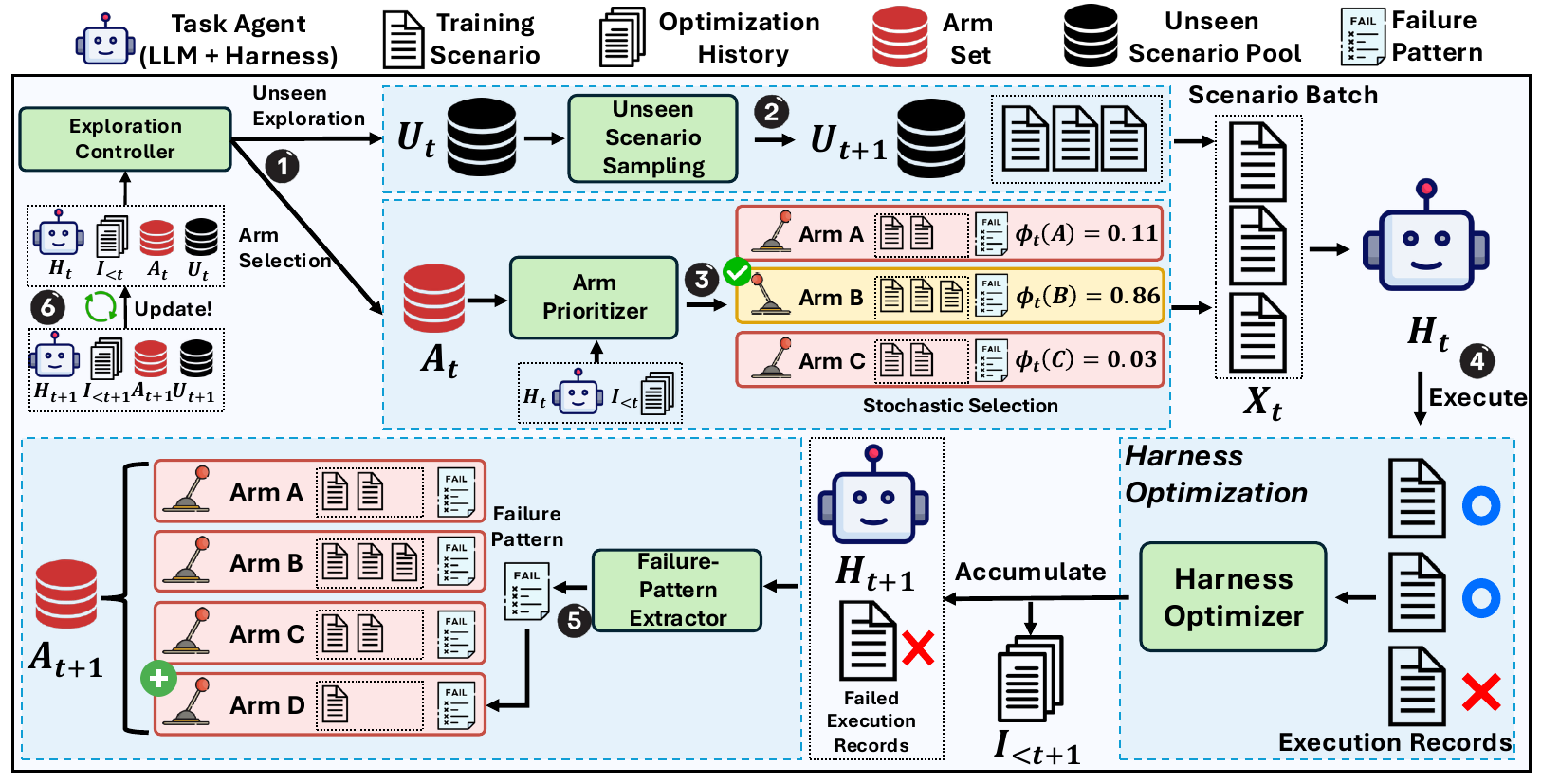}
    \vspace{-0.2cm}
    \caption{\textbf{Overview of \proposed}. At each iteration, the curriculum chooses between exploring unseen scenarios and revisiting an instantiated arm, optimizes the harness on the selected scenario batch, and updates its state based on the resulting execution and optimization outcomes.}
    \label{fig:activesaddler_overview}
\end{figure}

\paragraph{Workflow Overview.}

As illustrated in \autoref{fig:activesaddler_overview}, each iteration \(t\) begins with the current harness \(H_t\), accumulated optimization history \(\mathcal{I}_{<t}\), instantiated arm set \(\mathcal{A}_t\), and unseen scenario pool \(U_t\). The Exploration Controller first decides whether the next optimization step should explore previously unseen scenarios or continue optimizing a known failure-pattern arm (\bcirc{1}). If exploration is selected, \proposed{} constructs the next scenario batch \(X_t\) from \(U_t\) (\bcirc{2}); otherwise, the Arm Prioritizer selects an arm \(a_t\in\mathcal{A}_t\), and \(X_t\) is constructed from the scenarios associated with that arm (\bcirc{3}).

The selected scenarios are then executed under \(H_t\), and the resulting execution records \(\mathcal{E}_t\) are passed to the harness optimizer \(\mathcal{O}\) to produce the updated harness \(H_{t+1}\) (\bcirc{4}). From failed executions, the Failure-Pattern Extractor identifies recurring harness weaknesses, matches them to existing failure-pattern arms or instantiates new ones, and updates the arm set for subsequent iterations (\bcirc{5}). The resulting \(H_{t+1}\), optimization history \(\mathcal{I}_{<t+1}\), arm set \(\mathcal{A}_{t+1}\), and unseen scenario pool \(U_{t+1}\) define the state for the next curriculum decision (\bcirc{6}). In this way, the training curriculum co-evolves with the harness as its unresolved weaknesses change over optimization. The full ActiveSaddler algorithm and its instantiation with AutoSaddler are in~\autoref{app:activesaddler_algorithm}.

\paragraph{Failure-Pattern Extraction.}
\label{sec:arm_definition}

The Failure-Pattern Extractor converts failure evidence produced during each optimization step into reusable failure-pattern arms. Let $\mathcal{F}_t$ denote the failed execution records made available by the underlying harness optimizer at iteration $t$, and let $\delta_t(e)$ denote the associated diagnostic evidence for each $e\in\mathcal{F}_t$. The exact composition of $\mathcal{F}_t$ depends on the optimizer. In our AutoSaddler instantiation, it includes both failures diagnosed before patching and failures that remain or newly emerge during post-patch evaluation.

The extractor operates in two stages: \emph{symptom extraction} and \emph{pattern normalization}. It first independently abstracts each failure into a candidate symptom,
$c_t(e)=g_{\mathrm{ext}}(e,\delta_t(e))$,
without consulting the existing arm set. After all candidates are extracted, they are normalized against the current arms:
\begin{equation}
    \mathcal{A}_{t+1}
    =
    g_{\mathrm{norm}}\!\left(
        \mathcal{A}_t,
        \{c_t(e)\}_{e\in\mathcal{F}_t}
    \right),
    \label{eq:method_pattern_normalization}
\end{equation}
where $g_{\mathrm{norm}}$ links each candidate to an existing atomic pattern, composes multiple existing patterns when needed, or instantiates a new arm for a previously unseen failure pattern.

Each arm $a\in\mathcal{A}_{t+1}$ maintains a pattern description, a supporting scenario set $S_{t+1}(a)\subseteq D_{\mathrm{train}}$, and the associated harness, execution, and diagnostic evidence under which the pattern was observed. 
When an arm is selected, its supporting scenarios form the pool for the next optimization batch, subject to the per-iteration execution limit. Detailed extraction instructions are in \autoref{sec:instructions_used_in_activesaddler}.

\paragraph{Arm Prioritization.}
\label{sec:arm_selection}
When the Exploration Controller chooses to continue optimization over the
instantiated arm set, the Arm Prioritizer determines which
$a\in\mathcal{A}_t$ should receive the next optimization attempt.
Ideally, this decision would be based on the expected population-performance change
induced by selecting arm $a$.
However, this population objective is not directly observable during optimization; we therefore use an LLM to estimate a learning-progress score
\begin{equation}
    \phi_t(a)
    =
    g_{\mathrm{select}}(a,\mathcal{I}_{<t})
    \in [0,1],
    \label{eq:method_llm_arm_selection}
\end{equation}
where $g_{\mathrm{select}}$ denotes the LLM-based arm-scoring procedure and a larger
$\phi_t(a)$ indicates greater estimated benefit from allocating the next optimization
step to arm $a$.
To estimate this score, the LLM is instructed to jointly consider four factors: whether the failure pattern remains active under the current harness (\emph{severity}), whether it appears addressable through a harness intervention (\emph{fixability}), how broadly a successful repair may generalize (\emph{breadth}), and the risk that such an intervention introduces regressions (\emph{side-effect risk}).
The scores are recomputed from the latest $\mathcal{I}_{<t}$ at each arm-selection step, allowing arm priorities to adapt as the harness evolves.


The resulting scores define a stochastic selection distribution over the current arm set:
\begin{equation}
    q_t(a)
    =
    \frac{
        \exp\!\left(\phi_t(a)/\tau_{\mathrm{sel}}\right)
    }{
        \displaystyle
        \sum_{a'\in\mathcal{A}_t}
        \exp\!\left(\phi_t(a')/\tau_{\mathrm{sel}}\right)
    },
    \qquad
    a\in\mathcal{A}_t,
    \label{eq:method_arm_selection_distribution}
\end{equation}
where $\tau_{\mathrm{sel}}>0$ controls the stochasticity of arm selection.
The curriculum then samples
\begin{equation}
    a_t
    \sim
    \operatorname{Categorical}(q_t)
    \label{eq:method_selected_arm}
\end{equation}
and constructs the next scenario batch from the selected arm,
$X_t\subseteq S_t(a_t)$.
This stochastic selection rule favors arms with greater estimated learning progress while preserving the possibility of revisiting lower-scored arms as their utility changes over optimization.
Detailed scoring instructions and definitions of the four factors are in \autoref{sec:instructions_used_in_activesaddler}.

\paragraph{Exploration Control.}
\label{sec:unseen_exploration}
The Exploration Controller determines whether the next optimization step should exploit a previously discovered failure-pattern arm or explore unseen scenarios that may reveal new failure patterns.
Given the current harness $H_t$, instantiated arm set $\mathcal{A}_t$, unseen scenario pool $U_t$, and optimization history $\mathcal{I}_{<t}$, we instantiate an agent session for a binary decision:
\begin{equation}
    d_t
    =
    g_{\mathrm{explore}}
    \left(
        H_t,\mathcal{A}_t,U_t,\mathcal{I}_{<t}
    \right)
    \in
    \{\mathtt{unseen},\mathtt{arm}\}.
    \label{eq:method_llm_exploration_policy}
\end{equation}
The policy makes this decision based on the current optimization context, weighing whether the known arms contain a worthwhile unresolved optimization target against exploring the remaining unseen scenarios for previously undiscovered weaknesses. If $d_t=\mathtt{unseen}$, i.e., a \textsc{Draw} decision, \proposed{} samples $X_t\subseteq U_t$ without replacement and removes the selected scenarios from the unseen pool.
If $d_t=\mathtt{arm}$, i.e., a \textsc{Pull} decision, Arm Prioritizer selects $a_t\in\mathcal{A}_t$, and the next scenario batch is constructed as $X_t\subseteq S_t(a_t)$.
When $\mathcal{A}_t=\varnothing$ and $U_t\neq\varnothing$, the policy selects $\mathtt{unseen}$; when $U_t=\varnothing$ and $\mathcal{A}_t\neq\varnothing$, it selects $\mathtt{arm}$. 
Detailed decision criteria and instructions are in \autoref{sec:instructions_used_in_activesaddler}.

To complement the regression and generalization checks of the underlying harness optimizer, \proposed{} also guards against regressions on previously successful training scenarios. Once all offline training scenarios have been explored and every arm has since been visited, prior successes that did not instantiate an arm become eligible for exploration again, analogous to additional epochs over seen examples in conventional offline learning. If such a scenario now fails, the Failure-Pattern Extractor processes it normally, enabling regression rediscovery without re-evaluating all prior successes after every harness update.

\section{Experiments}
\label{sec:experiments}
\vspace{-0.2cm}
\subsection{Experimental Setup}
\label{sec:experimental_setup}
\vspace{-0.2cm}
\paragraph{Benchmarks and Base Harness Systems.}
We evaluate \proposed{} on two diverse benchmarks: GAIA2~\citep{GAIA2} and Terminal-Bench 2.0 (TB2)~\citep{TerminalBench2}. 
GAIA2 evaluates general-purpose assistant capabilities in a simulated smartphone environment spanning 10 distinct user environments, referred to as \textit{Universes}; we use its default ReAct-based agent as the base harness.
TB2 comprises 89 realistic tasks across domains including system administration, machine learning, and cybersecurity, and introduces Terminus 2, which we adopt as the base harness.

\paragraph{Implementation and Baseline Methods.}
We instantiate \proposed{} on AutoSaddler~\citep{autosaddler} by adding three curriculum components: the \textit{Failure-Pattern Extractor}, \textit{Arm Prioritizer}, and \textit{Exploration Controller} (Algorithm~\autoref{alg:activesaddler_autosaddler}).
These components use the GitHub Copilot SDK (GC-SDK)~\citep{CopilotSDK} and a shared pattern registry for dynamically instantiated failure-pattern arms; agent instructions and registry details are provided in \autoref{sec:instructions_used_in_activesaddler} and \autoref{app:more_details_exp_setup}.
We compare \proposed{} against GEPA~\citep{agrawal2026gepa}, Meta-Harness~\citep{lee2026meta}, and AutoSaddler~\citep{autosaddler}, which optimize prompts or harnesses without an explicit curriculum.
We also construct fixed easy-to-hard curricula~\citep{DBLP:conf/icml/BengioLCW09,DBLP:journals/corr/ZarembaS14,DBLP:conf/iclr/WuT17} by ranking categories or scenarios by empirical accuracy from five evaluations of the base harness on the training set.
For reference, we additionally report Terminus-KIRA~\citep{terminuskira2026}, a manually engineered TB2 harness.


\paragraph{Experimental Protocol and Data Splits.}
Unless otherwise noted, both the task agent and optimizer use \texttt{gpt-5.5}, with reasoning effort set to \texttt{medium} for the task agent and \texttt{xhigh} for GC-SDK optimizer sessions.
Following AutoSaddler~\citep{autosaddler}, we use out-of-distribution train/dev/test splits with disjoint task groups to evaluate harness generalization.
Detailed train/dev/test splits and hyperparameters are provided in \autoref{app:more_details_exp_setup}.
We evaluate each optimized harness over three test-time executions and report mean \textit{Pass@1} with standard deviation.


\subsection{Main Results}
\label{sec:main_results}
\vspace{-0.2cm}
\begin{table}[tbp]
\centering
\footnotesize
\setlength{\tabcolsep}{3pt}
\caption{Test Pass@1 on GAIA2 and Terminal-Bench 2.0 (mean $\pm$ std. over three test-time executions). Parentheses show task counts; bold indicates the best result.}
\vspace{-0.2cm}
\label{tab:eval_results_combined}
\sisetup{
  separate-uncertainty    = true,
  retain-zero-uncertainty = true,
  table-format            = 2.1(2),
  detect-weight           = true,
  detect-inline-weight    = math,
}

\begin{minipage}[t]{0.48\textwidth}
\centering
\vspace{0pt}
\textbf{GAIA2 (Pass@1)}
\vspace{0.25em}

\resizebox{\linewidth}{!}{%
\begin{tabular}{lS}
\toprule
\textbf{Harness (Type)}
 & {\textbf{Test-Split (300)}} \\
\midrule
Default Agent (manual)                       & {53.6 ± 1.1} \\
GEPA                                         & {54.2 ± 2.2} \\
Meta-Harness                                 & {54.2 ± 1.2} \\
AutoSaddler                                  & {55.4 ± 1.2} \\
AutoSaddler w/ Category Acc. Order           & {55.9 ± 1.3} \\
AutoSaddler w/ Scenario Acc. Order           & {55.7 ± 1.2} \\
\textbf{\proposed{}}                         & \bfseries {59.8 ± 1.0} \\
\midrule
w/o Failure Pattern Based Arm (Category Arm) & {56.8 ± 0.8} \\
w/o Failure Pattern Based Arm (Scenario Arm) & {56.2 ± 0.2} \\
w/o Arm Prioritizer                    & {55.3 ± 2.1} \\
w/o Exploration Controller    & {55.3 ± 0.9} \\
\bottomrule
\end{tabular}%
}
\end{minipage}
\hfill
\begin{minipage}[t]{0.48\textwidth}
\centering
\vspace{0pt}
\textbf{Terminal-Bench 2.0 (Pass@1)}
\vspace{0.25em}

\resizebox{\linewidth}{!}{%
\begin{tabular}{lS}
\toprule
\textbf{Harness (Type)}
 & {\textbf{Test-Split (40)}} \\
\midrule
Terminus 2 (manual)                          & {64.2 ± 2.9} \\
Terminus-KIRA (manual)                       & {69.2 ± 3.8} \\
GEPA                                         & {65.8 ± 5.2} \\
Meta-Harness                                 & {66.7 ± 5.2} \\
AutoSaddler                                  & {72.5 ± 0.0} \\
AutoSaddler w/ Category Acc. Order           & {70.8 ± 1.4} \\
AutoSaddler w/ Scenario Acc. Order           & {73.3 ± 1.4} \\
\textbf{\proposed{}}                         & \bfseries {80.0 ± 2.5} \\
\midrule
w/o Failure Pattern Based Arm (Category Arm) & {69.2 ± 2.9} \\
w/o Failure Pattern Based Arm (Scenario Arm) & {73.3 ± 3.8} \\
w/o Arm Prioritizer                    & {72.5 ± 2.5} \\
w/o Exploration Controller    & {71.7 ± 1.4} \\
\bottomrule
\end{tabular}%
}
\end{minipage}

\end{table}

As shown in \autoref{tab:eval_results_combined}, \proposed{} achieves the strongest test performance on both GAIA2 and TB2, reaching $59.8\%$ and $80.0\%$ Pass@1, respectively.
Under the same AutoSaddler optimizer, \proposed{} improves over a fixed randomly shuffled scenario order by \textbf{+4.4\,pp} on GAIA2 and \textbf{+7.5\,pp} on TB2.
It also outperforms difficulty-based fixed curricula by \textbf{+3.9\,pp} and \textbf{+4.1\,pp} over category- and scenario-level ordering on GAIA2, and by \textbf{+9.2\,pp} and \textbf{+6.7\,pp} on TB2, respectively.
Together, these results show that adapting scenario selection during optimization improves harness quality over fixed curricula under the same optimizer.

We further examine these gains through an independent optimization run, a different offline optimizer, and alternative curriculum strategies.
In an independent GAIA2 run, \proposed{} again outperforms all fixed-order baselines, showing robustness to optimization stochasticity (\autoref{app:optimization_stochasticity}).
Applying \proposed{} to GEPA improves its average Pass@1 from $54.2\%$ to $57.2\%$ (\textbf{+3.0\,pp}), showing that the benefit of adaptive training-scenario selection is not limited to AutoSaddler (\autoref{app:gepa_activesaddler}).
We also compare against non-LLM alternatives: EMA-based failure-persistence scoring and count-based UCB-AIR achieve results \textbf{3.1\,pp} and \textbf{4.2\,pp} below \proposed{}, respectively (\autoref{app:non_llm_alternatives}).

As shown in \autoref{fig:gaia2_curriculum_analysis}, \proposed{} achieves stronger development performance with fewer task-agent rollouts on both benchmarks.
We further analyze whether this efficiency persists after accounting for optimizer-side LLM overhead by combining it with task-agent execution cost.
Although \proposed{} incurs additional optimizer overhead, it achieves a substantially better end-to-end cost--accuracy trade-off on both benchmarks.
On GAIA2, \proposed{} reaches $58.5\%$ development accuracy at \$298, whereas AutoSaddler and category ordering require \$1{,}360 and \$673, respectively, to reach the same accuracy, and scenario ordering reaches only $56.9\%$ despite spending \$1{,}698.
On TB2, \proposed{} reaches $78.9\%$ at \$128, while AutoSaddler and scenario ordering require \$220 and \$363, respectively, to reach the same accuracy, and category ordering reaches only $73.7\%$ despite spending \$548.
Thus, the additional cost of curriculum adaptation is offset by more effective allocation of expensive task-agent rollouts; detailed cost analyses are provided in \autoref{app:full_optimization_cost}.

\subsection{Ablation Studies and Analysis}
\paragraph{RQ1: How do failure-pattern arms enable optimization to track and retarget weaknesses as the harness evolves?}
We first examine how the choice of arm representation affects optimization.
Specifically, we replace the failure-pattern arms in \proposed{} with category- or scenario-level arms while keeping the curriculum procedure unchanged.
As shown in \autoref{tab:eval_results_combined}, category and scenario arms reduce test Pass@1 by \textbf{3.0\,pp} and \textbf{3.6\,pp} on GAIA2, and by \textbf{10.8\,pp} and \textbf{6.7\,pp} on TB2, respectively.
These results show that arm representation is critical to harness quality.

Analyses in \autoref{app:more_details_RQ1} further support the benefit of failure-pattern arms.
The case studies show that failure-pattern arms preserve unresolved weaknesses as optimization targets, allowing them to be revisited and unsuccessful patches to be iteratively refined.
In contrast, category arms are too coarse, grouping distinct weaknesses together and allowing successes on some scenarios to mask unresolved weaknesses in others.
Scenario arms are too fine-grained, separating the same weakness across individual scenarios and making it harder to revisit under a limited budget.
Consistent with these cases, failure-pattern arms more frequently sample scenarios on which the current harness still fails, exceeding category and scenario arms by \textbf{16.5\,pp} and \textbf{35.3\,pp} on GAIA2, and by \textbf{29.9\,pp} and \textbf{23.8\,pp} on TB2, respectively.
The discovered failure pattern arms are listed in \autoref{app:failure_pattern_taxonomy}.

\paragraph{RQ2: How does adaptive arm scoring prioritize weaknesses as the harness evolves?}

\begin{figure}[!t]
    \centering
    \begin{subfigure}[b]{0.64\linewidth}
        \centering
        \includegraphics[width=\linewidth]{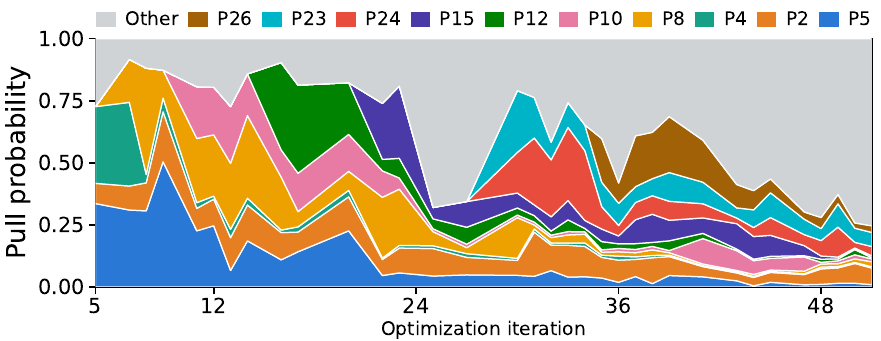}
        \caption{Evolution of pull-probability mass (GAIA2).}
        \label{fig:rq2_arm_priority_evolution}
    \end{subfigure}
    \hfill
    \begin{subfigure}[b]{0.34\linewidth}
        \centering
        \includegraphics[width=\linewidth]{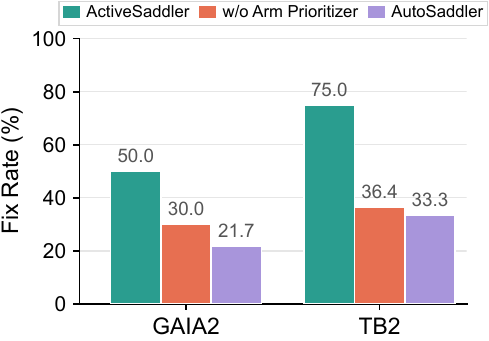}
        \caption{Failure-to-success conversion rate.}
        \label{fig:rq2_quantitative_analysis}
    \end{subfigure}
    \vspace{-0.2cm}
    \caption{
        \textbf{Effect of adaptive arm scoring.}
        Adaptive scoring dynamically reallocates optimization priority and yields more effective repair of observed failures.
    }
    \label{fig:rq2_arm_scoring}
\end{figure}

We examine whether prioritizing failure-pattern arms by their estimated learning potential improves optimization.
Specifically, we compare \proposed{} with a \textit{w/o Arm Prioritizer} variant that assigns the same score to every arm while keeping the remaining curriculum procedure unchanged.
As shown in \autoref{tab:eval_results_combined}, removing adaptive arm scoring reduces test Pass@1 by \textbf{4.5\,pp} on GAIA2 and \textbf{7.5\,pp} on TB2, showing the value of prioritizing by learning potential under limited rollouts.

To understand why adaptive arm scoring improves optimization, \autoref{fig:rq2_arm_scoring} examines both how optimization priorities evolve and how effectively observed failures are repaired.
As shown in (a), pull-probability mass continually shifts across failure-pattern arms rather than remaining concentrated on a fixed subset, indicating that the curriculum reallocates optimization budget as the harness and accumulated evidence evolve.
In (b), we measure the fraction of scenarios observed to fail during optimization that are subsequently converted to success by the optimized harness.
\proposed{} achieves the highest conversion rate, followed by \textit{w/o Arm Prioritizer} and AutoSaddler.
Further analyses in \autoref{app:more_details_RQ2} provide representative case studies of resolved and recurring weaknesses, examine probability redistribution across the full arm pool, and analyze stochastic selection as new arms are discovered.
Together, these results show that adaptive arm scoring improves optimization by reprioritizing weaknesses as their learning potential changes.

\paragraph{RQ3: How does adaptive exploration balance the continued repair of known weaknesses with the discovery of new ones?}

\begin{figure}[!t]
    \centering
    \includegraphics[width=\linewidth]{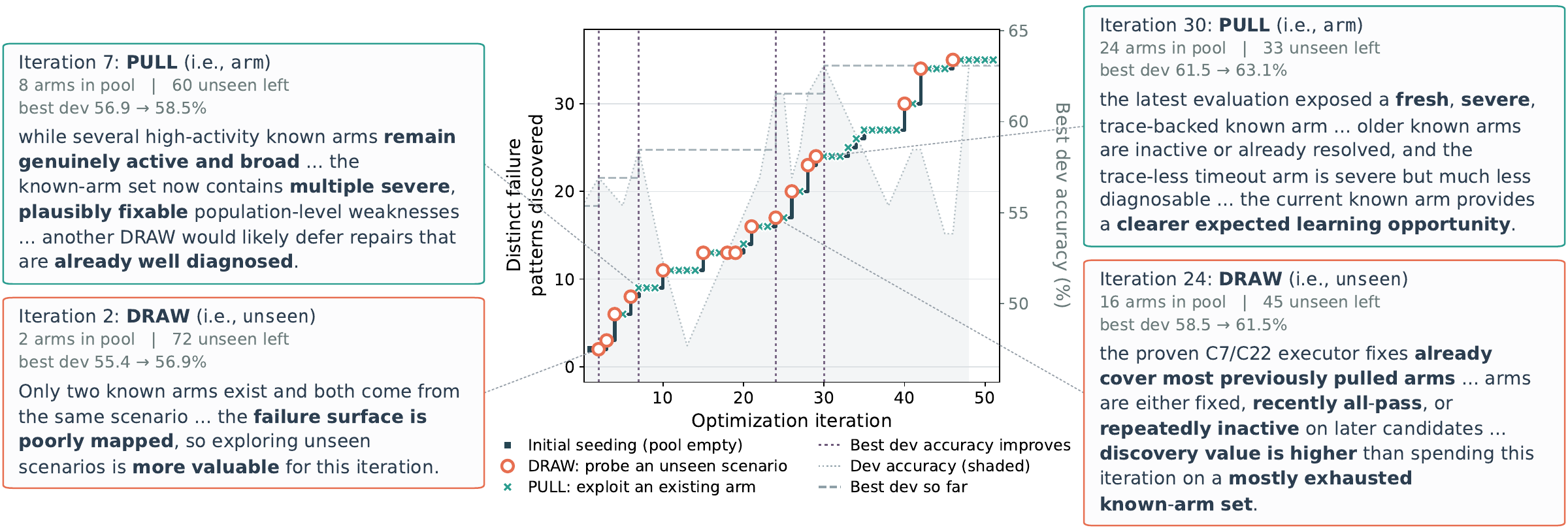}
    \caption{
    \textbf{Adaptive exploration over the optimization trajectory.}
    The policy chooses \textsc{Draw} when discovery is more valuable and \textsc{Pull} when promising repair opportunities remain.
    }
    \vspace{-0.2cm}
    \label{fig:rq3_adaptive_exploration}
\end{figure}

We examine whether adapting the timing of unseen-scenario exploration improves optimization over a fixed schedule.
Specifically, we compare \proposed{} with a \textit{w/o Exploration Controller} variant that explores unseen scenarios every five iterations while keeping the remaining curriculum procedure unchanged.
As shown in \autoref{tab:eval_results_combined}, replacing adaptive exploration with this fixed schedule reduces test Pass@1 by \textbf{4.5\,pp} on GAIA2 and \textbf{8.3\,pp} on TB2.
These results show that when to explore is critical under a limited optimization budget.

\autoref{fig:rq3_adaptive_exploration} illustrates how exploration decisions adapt to the evolving optimization state on GAIA2.
Representative decisions show that the policy selects \textsc{Draw} when known arms offer limited remaining learning value and \textsc{Pull} when actionable weaknesses remain.
This behavior also appears globally: the \textsc{Draw} rate decreases from 44\% in the first half, with relatively few instantiated arms, to 20\% in the second half, with a larger arm pool.
To quantify this behavior, we compare how many arms remain unresolved according to their most recent observation at \textsc{Pull} and \textsc{Draw} decisions.
As detailed in \autoref{app:more_details_RQ3}, \proposed{} has 2.0$\times$ and 6.9$\times$ as many unresolved arms at \textsc{Pull} as at \textsc{Draw} decisions on GAIA2 and TB2, respectively.
In contrast, the fixed schedule yields ratios close to one (0.8$\times$ and 1.3$\times$), indicating that its \textsc{Pull} and \textsc{Draw} decisions are largely insensitive to how many unresolved arms remain.
Together, these results show that adaptive exploration favors repair when worthwhile unresolved weaknesses remain and discovery when few such weaknesses remain among the known arms.
\section{Conclusion}
\label{sec:conclusion}
\vspace{-0.2cm}
We introduced \proposed{}, an automated curriculum-learning approach for harness optimization that adapts which training scenarios generate the execution feedback used for each harness update. 
Rather than fixing scenario selection before optimization, \proposed{} formulates curriculum construction as a non-stationary bandit that dynamically instantiates failure-pattern arms, prioritizes them by their evolving learning potential, and balances revisiting known weaknesses with exploring unseen scenarios. 
Across GAIA2 and Terminal-Bench 2.0, \proposed{} improves test Pass@1 by \textbf{4.4} and \textbf{7.5} percentage points over the same harness optimizer with a fixed scenario order, respectively, while also outperforming existing harness optimizers and fixed curricula. 
Our ablations further show that these gains rely on failure-pattern arms, adaptive arm prioritization, and adaptive exploration. 
Overall, these results establish automated curriculum learning as a complementary dimension of harness optimization: performance depends not only on how feedback is used for updates, but also on which training scenarios generate that feedback.

\subsection*{AI Use Statement}
\label{sec:ai_use_statement}
We used generative AI tools for writing assistance, including revising sentences and paragraphs to improve clarity, readability, and overall presentation.
We also used generative AI tools for retrieval and discovery, including identifying potentially relevant related work and supporting literature searches.
All AI-assisted content was carefully reviewed and verified by the authors, including the wording of the manuscript, technical claims, numerical results, and citations.
The authors take full responsibility for the final manuscript.

\subsection*{Reproducibility Statement}
\label{sec:reproducibility_statement}
We provide the implementation details, prompts, data splits, and evaluation protocol needed to reproduce our experiments.
Section~\ref{sec:experimental_setup} specifies the benchmark and base-harness configurations, baseline methods, model settings, construction of the fixed curricula, and test-time evaluation protocol.
Appendix~\ref{app:more_details_exp_setup} further describes the Pattern Registry and its \texttt{pattern} CLI interface, as well as the data-split construction and task counts for both benchmarks.
Algorithm~\ref{alg:activesaddler_general} provides the complete optimization procedure, including curriculum decisions, and development-set-based harness selection.
Finally, Appendix~\ref{sec:instructions_used_in_activesaddler} provides the full session prompt templates used for failure-pattern extraction, arm scoring, and unseen-scenario exploration.

\bibliography{iclr2027_conference}
\bibliographystyle{iclr2027_conference}

\newpage
\appendix

\section{Additional Details on Experiment Setup}
\label{app:more_details_exp_setup}

\paragraph{Implementation of Pattern Registry.}
\label{para:pattern_cli}
The Pattern Registry stores each failure pattern, i.e., each arm, as a symptom-level label with its tagged \emph{(harness, trace, scenario)} tuples, root-cause evidence, activity observations, and learning-progress scores. 
The registry can be implemented straightforwardly, but providing effective agent access to it is more challenging: as arms accumulate observation histories and evidence across iterations, serializing the registry into the prompt is neither compact nor selective. 
We therefore introduce the \texttt{pattern} command-line interface (CLI) to facilitate interaction between the GC-SDK and the Pattern Registry. 
The interface provides on-demand access to scoped views of the arm population, including the canonical pattern table, per-pattern evidence, score rankings, and the complete pull history of an arm with its patch intents; it is also the only channel through which the agent registers arms, tags tuples, and records scores and decisions. 
Because sessions differ in what they are permitted to do, the outer loop installs the CLI with a per-session capability manifest determined by the sampling strategy, and commands outside it are refused. 
The full command set is summarized in~\autoref{tab:pattern_commands}.

\begin{table}[h]
\centering
\footnotesize
\caption{Commands exposed by the \texttt{pattern} CLI, grouped into read operations for scoped views of the arm population and write operations for per-session registry updates. Each session receives only the subset of commands in its capability manifest.}
\label{tab:pattern_commands}
\small
\begin{tabular}{@{}p{0.32\linewidth} p{0.32\linewidth} p{0.30\linewidth}@{}}
\toprule
\textbf{Command} & \textbf{Purpose} & \textbf{When to use} \\
\midrule
\multicolumn{3}{@{}l}{\textit{Read operations}} \\
\midrule
\texttt{pattern list}
& Canonical arm table: score, observation history, scenario count, last observation, label
& Start of session; orienting to the full arm population \\
\addlinespace
\texttt{pattern show <id>}
& Pattern details: label, score, tagged tuples, per-scenario root causes, observations
& Inspecting a specific arm before rating or patching \\
\addlinespace
\texttt{pattern score [--top-k <n>]}
& Top-$k$ arms ranked by current score
& Quick triage of the highest-priority arms \\
\addlinespace
\texttt{pattern scenarios --pattern-id <id>}
& Scenarios tagged to an arm, with co-occurring patterns
& Checking an arm's coverage and overlap \\
\addlinespace
\texttt{pattern history [--pattern-id <ids>] [--last-k <K>]}
& Complete per-arm pull history: patched attempts, all-pass skips, and failed attempts, with diagnoses, patch intents, dev-set impact, and per-scenario reflections
& Before scoring or re-selecting an arm; assessing what prior repairs already attempted \\
\midrule
\multicolumn{3}{@{}l}{\textit{Write operations}} \\
\midrule
\texttt{pattern register --label "..."}
& Create a new arm from a symptom-level label; returns its pattern ID
& Pattern Extraction Session, for a newly observed failure type \\
\addlinespace
\texttt{pattern tag --pattern-id <ids> --harness <idx> --trace <dir> --scenario <sid> [--root-cause "..."]}
& Bind a \emph{(harness, trace, scenario)} tuple to one or more patterns and record its root-cause evidence
& Pattern Extraction Session, after normalizing a candidate against existing arms \\
\addlinespace
\texttt{pattern rate --pattern-id <id> --severity --fixability --breadth --side-effect [--rationale]}
& Record the cardinal learning-progress score for the current iteration from the severity, fixability, breadth, and side-effect axes
& Arm Scoring Session, once per arm; an unrated arm defaults to the lowest priority \\
\addlinespace
\texttt{pattern decide --action pull|draw [--rationale]}
& Record the exploration decision: exploit a known arm (\texttt{pull}) or explore the unseen-scenario pool for a new failure type (\texttt{draw})
& Unseen-Scenario Exploration Session, exactly once per iteration \\
\bottomrule
\end{tabular}
\end{table}

\paragraph{Data Splits across Benchmarks.}
To evaluate whether optimized harnesses generalize beyond the tasks seen during optimization, we use the same train, development, and test splits as AutoSaddler~\citep{autosaddler} for each benchmark, as summarized in~\autoref{tab:data_splits}. For benchmarks with a natural grouping structure, we split by task group rather than by individual tasks. In GAIA2, the split axis is the \texttt{Universe}, where each \texttt{Universe} corresponds to a distinct persona and simulated digital environment. This design ensures that the test set contains tasks from groups unseen during optimization, providing a stronger measure of out-of-distribution generalization than random task-level splits. For Terminal-Bench~2.0, tasks span diverse domains but do not provide a natural grouping axis; we therefore use a uniform random partition into train, dev, and test sets.

\begin{table}[htbp]
\centering
\footnotesize
\setlength{\tabcolsep}{5pt}
\renewcommand{\arraystretch}{1.1}
\caption{Data splits across the two benchmarks. For GAIA2, we evaluate generalization under distribution shift by partitioning the benchmark such that train, development, and test sets contain tasks from disjoint \texttt{Universes} (personas), rather than using random task-level splits. Terminal-Bench~2.0 contains only 89 tasks across diverse domains and offers no natural grouping axis, so we adopt a uniform random partition.}
\label{tab:data_splits}
\begin{tabular}{@{}llllr@{}}
\toprule
\textbf{Benchmark} & \textbf{Split Axis} & \textbf{Split} & \textbf{Task Group} & \textbf{\# Tasks} \\
\midrule
\multirow{5}{*}{\textbf{GAIA2}}
& \multirow{5}{*}{\texttt{Universe} (persona)}
& Train & \texttt{Universe 29} & 75 \\
\cmidrule(l){3-5}
& & Dev & \texttt{Universe 30} & 65 \\
\cmidrule(l){3-5}
& & \multirow{3}{*}{Test}
& \texttt{Universe 21} & 107 \\
& & & \texttt{Universe 22} & 112 \\
& & & \texttt{Universe 27} & 81 \\
\midrule
\multirow{3}{*}{\textbf{Terminal-Bench~2.0}}
& \multirow{3}{*}{Random$^{\dagger}$}
& Train & --- & 30 \\
\cmidrule(l){3-5}
& & Dev & --- & 19 \\
\cmidrule(l){3-5}
& & Test & --- & 40 \\
\bottomrule
\end{tabular}

\par\smallskip
\raggedright\scriptsize
$^{\dagger}$ Only 89 tasks across heterogeneous domains (system administration, machine learning, and cybersecurity); no natural axis for distribution-shifted splits exists.
\end{table}

\paragraph{Hyperparameters.}
We set the shared rollout budget to $K=10\,(|\mathcal{D}_{\mathrm{train}}|+|\mathcal{D}_{\mathrm{dev}}|)$, corresponding to 1,400 rollouts on GAIA2 and 490 on Terminal-Bench~2.0.
We choose this budget to match ten full passes over the Meta-Harness training set: following the configuration used in AutoSaddler~\citep{autosaddler}, Meta-Harness treats $\mathcal{D}_{\mathrm{train}} \cup \mathcal{D}_{\mathrm{dev}}$ as its training set, and Meta-Harness~\citep{lee2026meta} runs optimization for 10 iterations.
For all methods, we limit each optimization iteration to at most three scenarios, matching the mini-batch size used in AutoSaddler~\citep{autosaddler}.
We set the arm-selection temperature to $\tau_{\mathrm{sel}}=0.15$.

\section{Algorithm of \proposed}
\label{app:activesaddler_algorithm}

\paragraph{General Algorithm.}
Algorithm~\ref{alg:activesaddler_general} operationalizes the curriculum policy defined in Sections~\ref{sec:preliminaries} and~\ref{sec:method}.
At each iteration, the \textbf{Exploration Controller} first decides whether to explore scenarios from the current exploration pool or continue optimization of an instantiated failure-pattern arm.
If an arm is chosen, the \textbf{Arm Prioritizer} scores the current arms and samples the next optimization target.
After the fixed harness optimizer $\mathcal{O}$ updates the harness, the \textbf{Failure-Pattern Extractor} converts the resulting failure evidence into matched, composed, or newly instantiated failure-pattern arms.
Thus, the curriculum determines which training scenarios form $X_t$, while $\mathcal{O}$ determines how the resulting execution evidence is translated into harness updates and uses $D_{\mathrm{dev}}$ for candidate selection, regression checks, and generalization checks.

The regression-rediscovery rule in Section~\ref{sec:unseen_exploration} is handled within the Exploration Controller.
Once all training scenarios have been explored and every arm has since been visited, previously successful scenarios that did not instantiate an arm become eligible for exploration again.
This changes only the eligibility of scenarios for the $\mathtt{unseen}$ branch and does not introduce an additional curriculum action.
Following AutoSaddler~\citep{autosaddler}, all task-level executions incurred during optimization, including those used by the underlying optimizer to evaluate candidate harnesses, count toward the fixed rollout budget $K$, and the final harness is selected based on development-set performance.

\begin{algorithm}[t]
\caption{\proposed{} for automated curriculum learning in harness optimization}
\label{alg:activesaddler_general}
\small
\begin{algorithmic}[1]
\Require Initial harness parameters $\theta_0$, training set
$D_{\mathrm{train}}$, development set $D_{\mathrm{dev}}$, rollout budget $K$,
fixed harness optimizer $\mathcal{O}$, arm-selection temperature
$\tau_{\mathrm{sel}}$, and procedures
$g_{\mathrm{explore}},g_{\mathrm{select}},g_{\mathrm{ext}},g_{\mathrm{norm}}$
\State $H_0\gets H_{\theta_0}$
\State Initialize optimization history $\mathcal{I}_{<0}$ with $H_0$ and arm set
$\mathcal{A}_0\gets\varnothing$
\State $U_0\gets D_{\mathrm{train}}$; $t\gets 0$

\While{rollout budget remains}

    \Statex \textbf{Exploration Controller}
    \If{$U_t=\varnothing$, all scenarios in $D_{\mathrm{train}}$ have been
    explored, and every instantiated arm has since been visited}
        \State Add to $U_t$ previously successful scenarios that did not
        instantiate a failure-pattern arm
    \EndIf

    \If{$\mathcal{A}_t=\varnothing$ and $U_t\neq\varnothing$}
        \State $d_t\gets\mathtt{unseen}$
    \ElsIf{$U_t=\varnothing$ and $\mathcal{A}_t\neq\varnothing$}
        \State $d_t\gets\mathtt{arm}$
    \ElsIf{$\mathcal{A}_t\neq\varnothing$ and $U_t\neq\varnothing$}
        \State $d_t\gets
        g_{\mathrm{explore}}
        (H_t,\mathcal{A}_t,U_t,\mathcal{I}_{<t})$
    \Else
        \State \textbf{break}
    \EndIf

    \If{$d_t=\mathtt{unseen}$}
        \State Sample $X_t\subseteq U_t$ without replacement, subject to the
        per-iteration execution limit
        \State $U_t\gets U_t\setminus X_t$

    \Else
        \Statex \textbf{Arm Prioritizer}
        \ForAll{$a\in\mathcal{A}_t$}
            \State $\phi_t(a)\gets
            g_{\mathrm{select}}(a,\mathcal{I}_{<t})$
        \EndFor
        \State $q_t(a)\gets
        \dfrac{\exp(\phi_t(a)/\tau_{\mathrm{sel}})}
        {\sum_{a'\in\mathcal{A}_t}
        \exp(\phi_t(a')/\tau_{\mathrm{sel}})}$
        for all $a\in\mathcal{A}_t$
        \State $a_t\sim\operatorname{Categorical}(q_t)$
        \State Construct $X_t\subseteq S_t(a_t)$, subject to the
        per-iteration execution limit
    \EndIf

    \Statex \textbf{Harness Optimizer}
    \State $\mathcal{E}_t\gets\Call{Execute}{H_t,X_t}$
    \State $\theta_{t+1}\sim
    \mathcal{O}(\theta_t,\mathcal{E}_t,D_{\mathrm{dev}})$;
    $H_{t+1}\gets H_{\theta_{t+1}}$
    \State Obtain failed execution records $\mathcal{F}_t$ and diagnostic
    evidence $\{\delta_t(e)\}_{e\in\mathcal{F}_t}$ made available by
    $\mathcal{O}$

    \Statex \textbf{Failure-Pattern Extractor}
    \ForAll{$e\in\mathcal{F}_t$}
        \State $c_t(e)\gets g_{\mathrm{ext}}(e,\delta_t(e))$
        \Comment{Symptom extraction}
    \EndFor
    \State $\mathcal{A}_{t+1}\gets
    g_{\mathrm{norm}}\!\left(
        \mathcal{A}_t,
        \{c_t(e)\}_{e\in\mathcal{F}_t}
    \right)$
    \Comment{Pattern normalization}
    \State Update the supporting scenario sets and associated harness,
    execution, and diagnostic evidence for matched, composed, or newly instantiated arms

    \Statex \textbf{Curriculum-State Update}
    \State $U_{t+1}\gets U_t$
    \State Update $\mathcal{I}_{<t+1}$ with $H_{t+1}$, $X_t$, the execution
    and optimizer outcomes, and the curriculum decision
    \State $t\gets t+1$
\EndWhile

\State Select $\hat{\theta}_\pi$ by development-set performance
\State \Return $H_{\hat{\theta}_\pi}$
\end{algorithmic}
\end{algorithm}
\newpage

\begin{algorithm}[!htbp]
\caption{\proposed{} instantiated on AutoSaddler
(\ASdiff{AutoSaddler-specific operations highlighted in blue})}
\label{alg:activesaddler_autosaddler}
\small
\begin{algorithmic}[1]
\Require Initial harness parameters $\theta_0$, training set
$D_{\mathrm{train}}$, development set $D_{\mathrm{dev}}$, rollout budget $K$,
\ASdiff{AutoSaddler as the fixed harness optimizer $\mathcal{O}$},
arm-selection temperature $\tau_{\mathrm{sel}}$, and procedures
$g_{\mathrm{explore}},g_{\mathrm{select}},g_{\mathrm{ext}},g_{\mathrm{norm}}$
\State $H_0\gets H_{\theta_0}$
\State Initialize optimization history $\mathcal{I}_{<0}$ with $H_0$ and arm set
$\mathcal{A}_0\gets\varnothing$
\State $U_0\gets D_{\mathrm{train}}$; $t\gets 0$

\While{rollout budget remains}

    \Statex \textbf{Exploration Controller}
    \If{$U_t=\varnothing$, all scenarios in $D_{\mathrm{train}}$ have been
    explored, and every instantiated arm has since been visited}
        \State Add to $U_t$ previously successful scenarios that did not
        instantiate a failure-pattern arm
    \EndIf

    \If{$\mathcal{A}_t=\varnothing$ and $U_t\neq\varnothing$}
        \State $d_t\gets\mathtt{unseen}$
    \ElsIf{$U_t=\varnothing$ and $\mathcal{A}_t\neq\varnothing$}
        \State $d_t\gets\mathtt{arm}$
    \ElsIf{$\mathcal{A}_t\neq\varnothing$ and $U_t\neq\varnothing$}
        \State $d_t\gets
        g_{\mathrm{explore}}
        (H_t,\mathcal{A}_t,U_t,\mathcal{I}_{<t})$
    \Else
        \State \textbf{break}
    \EndIf

    \If{$d_t=\mathtt{unseen}$}
        \State Sample $X_t\subseteq U_t$ without replacement, subject to the
        per-iteration execution limit
        \State $U_t\gets U_t\setminus X_t$

    \Else
        \Statex \textbf{Arm Prioritizer}
        \ForAll{$a\in\mathcal{A}_t$}
            \State $\phi_t(a)\gets
            g_{\mathrm{select}}(a,\mathcal{I}_{<t})$
        \EndFor
        \State $q_t(a)\gets
        \dfrac{\exp(\phi_t(a)/\tau_{\mathrm{sel}})}
        {\sum_{a'\in\mathcal{A}_t}
        \exp(\phi_t(a')/\tau_{\mathrm{sel}})}$
        for all $a\in\mathcal{A}_t$
        \State $a_t\sim\operatorname{Categorical}(q_t)$
        \State Construct $X_t\subseteq S_t(a_t)$, subject to the
        per-iteration execution limit
    \EndIf

    \Statex \ASdiff{\textbf{AutoSaddler Optimizer}}
    \State $\mathcal{E}_t\gets\Call{Execute}{H_t,X_t}$
    \If{$\Call{Failed}{\mathcal{E}_t}\neq\varnothing$}
        \State \ASdiff{Diagnose failures in $\mathcal{E}_t$ and derive a
        structured patch, yielding candidate harness $H'_t$}
        \State \ASdiff{$\mathcal{E}^{\mathrm{post}}_t\gets
        \Call{Execute}{H'_t,X_t}$}
        \State \ASdiff{If $H'_t$ improves over $H_t$ on $X_t$, additionally
        evaluate $H'_t$ on $D_{\mathrm{dev}}$}
        \State \ASdiff{Reflect on the pre-/post-patch outcomes; record
        \emph{fixed}, \emph{regressed}, \emph{still-failing}, and
        \emph{still-passing} cases; and update EvoDAG}
        \State \ASdiff{$H_{t+1}\gets\Call{Evolve}{\mathrm{EvoDAG}}$}
        \State \ASdiff{$\mathcal{F}_t\gets
        \Call{Failed}{\mathcal{E}_t}\cup
        \Call{Failed}{\mathcal{E}^{\mathrm{post}}_t}$, with diagnostic evidence
        $\{\delta_t(e)\}_{e\in\mathcal{F}_t}$ from the Diagnosis--Patch and
        Reflection sessions}
    \Else
        \State \ASdiff{$H_{t+1}\gets H_t$; $\mathcal{F}_t\gets\varnothing$}
    \EndIf

    \Statex \textbf{Failure-Pattern Extractor}
    \ForAll{$e\in\mathcal{F}_t$}
        \State $c_t(e)\gets g_{\mathrm{ext}}(e,\delta_t(e))$
        \Comment{Symptom extraction}
    \EndFor
    \State $\mathcal{A}_{t+1}\gets
    g_{\mathrm{norm}}\!\left(
        \mathcal{A}_t,
        \{c_t(e)\}_{e\in\mathcal{F}_t}
    \right)$
    \Comment{Pattern normalization}
    \State Update the supporting scenario sets and associated harness,
    execution, and diagnostic evidence for matched, composed, or newly instantiated arms

    \Statex \textbf{Curriculum-State Update}
    \State $U_{t+1}\gets U_t$
    \State Update $\mathcal{I}_{<t+1}$ with $H_{t+1}$, $X_t$, the execution
    and optimizer outcomes, and the curriculum decision
    \State $t\gets t+1$
\EndWhile

\State Select $\hat{\theta}_\pi$ by development-set performance
\State \Return $H_{\hat{\theta}_\pi}$
\end{algorithmic}
\end{algorithm}
\newpage

\paragraph{AutoSaddler Instantiation.}
Algorithm~\ref{alg:activesaddler_autosaddler} instantiates the same
\proposed{} curriculum on AutoSaddler.
The three curriculum components, the \textbf{Exploration Controller},
\textbf{Arm Prioritizer}, and \textbf{Failure-Pattern Extractor}, are
unchanged from Algorithm~\ref{alg:activesaddler_general}.
We repeat the full procedure for line-by-line comparison, with only the
AutoSaddler-specific optimizer operations highlighted in blue.

The AutoSaddler-specific difference lies in how the selected execution records
are converted into a harness update and which failures are exposed to the
Failure-Pattern Extractor.
After executing $H_t$ on $X_t$, the Diagnosis--Patch Session diagnoses the
observed failures and derives a structured patch, yielding a candidate harness
$H'_t$.
The candidate is then executed on the same scenario batch $X_t$ and, if it
improves over $H_t$ on this batch, is additionally evaluated on
$D_{\mathrm{dev}}$.
The Reflection Session compares the pre- and post-patch executions, records
\emph{fixed}, \emph{regressed}, \emph{still-failing}, and
\emph{still-passing} cases, and updates EvoDAG.
The Evolution Session then synthesizes $H_{t+1}$ from the updated EvoDAG.
For this instantiation, $\mathcal{F}_t$ contains failures diagnosed before
patching together with failures that remain or newly emerge during post-patch
evaluation, with diagnostic evidence supplied by the Diagnosis--Patch and
Reflection sessions.
The same Failure-Pattern Extractor then processes $\mathcal{F}_t$ exactly as
in the general algorithm.

\section{Robustness to Optimization Stochasticity}
\label{app:optimization_stochasticity}

\begin{table}[!h]
\centering
\small
\setlength{\tabcolsep}{5pt}
\caption{\textbf{Robustness to optimization stochasticity.}
Pass@1 on three held-out GAIA2 universes for harnesses obtained from two
independent optimization runs of AutoSaddler, its difficulty-based fixed
curricula, and \proposed{}.
Performance on each universe is reported as mean $\pm$ standard deviation over
three test-time executions, and the average is computed across all test scenarios.}
\label{tab:optimization_stochasticity}
\sisetup{
  separate-uncertainty = true,
  table-format         = 2.1(2),
  detect-weight        = true,
  detect-inline-weight = math,
}
\begin{tabular}{lSSSS}
\toprule
\multicolumn{1}{c}{\multirow{2}{*}{\textbf{Harness (Type)}}}
 & \multicolumn{3}{c}{\textbf{GAIA2 Universe (Pass@1)}}
 & \multicolumn{1}{c}{\multirow{2}{*}{\textbf{Avg.}}} \\
\cmidrule(lr){2-4}
 & {\textbf{21 (107)}} & {\textbf{22 (112)}} & {\textbf{27 (81)}} & \\
\midrule
Default Agent (manual)                               & {54.5 ± 0.5} & {50.0 ± 1.5} & {57.2 ± 1.4} & {53.6 ± 1.1} \\
AutoSaddler (Run 1)                                  & {57.0 ± 0.9} & {51.5 ± 2.2} & {58.8 ± 0.7} & {55.4 ± 1.2} \\
AutoSaddler (Run 2)                                  & {57.6 ± 1.4} & {53.0 ± 1.0} & {60.1 ± 2.9} & {56.6 ± 1.0} \\
AutoSaddler w/ Category Acc. Order (Run 1)           & {56.1 ± 3.4} & {53.9 ± 1.0} & {58.4 ± 1.9} & {55.9 ± 1.3} \\
AutoSaddler w/ Category Acc. Order (Run 2)           & {57.0 ± 3.4} & {53.6 ± 0.9} & {58.4 ± 0.7} & {56.1 ± 1.0} \\
AutoSaddler w/ Scenario Acc. Order (Run 1)           & {57.0 ± 1.9} & {53.3 ± 1.0} & {57.2 ± 3.8} & {55.7 ± 1.2} \\
AutoSaddler w/ Scenario Acc. Order (Run 2)           & {56.1 ± 3.7} & {52.7 ± 2.4} & {60.1 ± 1.9} & {55.9 ± 0.7} \\
\proposed{} (Run 1)                                  & {60.4 ± 1.4} & {57.7 ± 2.6} & {61.7 ± 1.2} & {59.8 ± 1.0} \\
\proposed{} (Run 2)                                  & \bfseries {62.0 ± 4.4} & \bfseries {60.4 ± 1.9} & \bfseries {62.6 ± 2.6} & \bfseries {61.6 ± 2.9} \\
\bottomrule
\end{tabular}
\end{table}

To assess robustness to optimization stochasticity, we conduct two independent GAIA2 optimization runs for AutoSaddler, its category- and scenario-level difficulty-based fixed curricula, and \proposed{}.
Each resulting harness is evaluated on three held-out universes (Universes~21, 22, and~27; 300 scenarios in total) with three repeated test-time executions.
Both runs of \proposed{} outperform all fixed-order baselines on each held-out universe.
Across the three universes, \proposed{} achieves 59.8\% and 61.6\% Pass@1 in Runs~1 and~2, exceeding the fixed-order baselines by 3.9 to 4.4 and 5.0 to 5.7 percentage points, respectively.
These consistent gains across independent optimization runs and held-out universes indicate that the improvement of \proposed{} is not specific to a single optimization trajectory.

\section{Alternative Arm Prioritization and Exploration Strategies}
\label{app:non_llm_alternatives}

\begin{table}[!h]
\centering
\small
\setlength{\tabcolsep}{5pt}
\caption{\textbf{Comparison with alternative curriculum strategies.}
Pass@1 on three held-out GAIA2 universes when replacing either the Arm
Prioritizer or Exploration Controller in \proposed{} with an alternative
strategy while keeping the remaining curriculum components unchanged.
Performance on each universe is reported as mean $\pm$ standard deviation over
three test-time executions.}
\label{tab:non_llm_alternatives}

\sisetup{
  separate-uncertainty = true,
  table-format         = 2.1(2),
  detect-weight        = true,
  detect-inline-weight = math,
}

\begin{tabular}{lSSSS}
\toprule
\multicolumn{1}{c}{\multirow{2}{*}{\textbf{Method}}}
 & \multicolumn{3}{c}{\textbf{GAIA2 Universe (Pass@1)}}
 & \multicolumn{1}{c}{\multirow{2}{*}{\textbf{Avg.}}} \\
\cmidrule(lr){2-4}
 & {\textbf{21 (107)}} & {\textbf{22 (112)}} & {\textbf{27 (81)}} & \\
\midrule
Default Agent (manual)                               & {54.5 ± 0.5} & {50.0 ± 1.5} & {57.2 ± 1.4} & {53.6 ± 1.1} \\
AutoSaddler
    & {57.0 ± 0.9}
    & {51.5 ± 2.2}
    & {58.8 ± 0.7}
    & {55.4 ± 1.2} \\
\proposed{}
    & \bfseries {60.4 ± 1.4}
    & \bfseries {57.7 ± 2.6}
    & \bfseries {61.7 ± 1.2}
    & \bfseries {59.8 ± 1.0} \\
\proposed{} w/ EMA Scoring
    & {57.9 ± 2.5}
    & {53.0 ± 3.4}
    & {60.1 ± 0.7}
    & {56.7 ± 1.5} \\
\proposed{} w/ UCB-AIR Controller
    & {57.9 ± 3.4}
    & {50.9 ± 1.5}
    & {58.8 ± 3.1}
    & {55.6 ± 1.4} \\
\midrule
w/o Arm Prioritizer
    & {57.3 ± 1.4}
    & {51.2 ± 3.1}
    & {58.4 ± 4.3}
    & {55.3 ± 2.1} \\
w/o Exploration Controller
    & {56.4 ± 2.2}
    & {51.5 ± 0.5}
    & {59.3 ± 3.3}
    & {55.3 ± 0.9} \\
\bottomrule
\end{tabular}
\end{table}

To examine alternative design choices for the two curriculum decisions in \proposed{}, we replace the Arm Prioritizer and Exploration Controller, one at a time, with simple non-LLM alternatives while keeping the Failure-Pattern
Extractor, arm-selection distribution, and underlying harness optimizer unchanged.

First, \textbf{\proposed{} w/ EMA Scoring} replaces the LLM-based learning-progress estimate $\phi_t(a)=g_{\mathrm{select}}(a,\mathcal{I}_{<t})$ with an exponential moving-average (EMA) score, following the EMA-based curriculum scoring used by TSCL~\citep{DBLP:journals/corr/MatiisenOCS17}.
In our setting, the score tracks the residual failure rate of each arm after patching:

$$
\phi_t(a)=\bar{X}_t(a), \qquad
\bar{X}_t(a)=
\begin{cases}
(1-\eta)\bar{X}_{t-1}(a)
+\eta\,\mathrm{active}_t(a),
& \text{if $a$ is observed at $t$},\\
\bar{X}_{t-1}(a),
& \text{otherwise},
\end{cases}
$$

where $\mathrm{active}_t(a)$ is the fraction of evaluated scenarios associated with $a$ that still exhibit the same failure pattern after patching.
Thus, recurring failures that remain unresolved receive higher scores and are prioritized for further optimization.
Following TSCL~\citep{DBLP:journals/corr/MatiisenOCS17}, we set $\eta=0.1$. 
Newly instantiated arms are initialized with $\bar{X}_t(a)=1$.

Second, \textbf{\proposed{} w/ UCB-AIR Controller} retains the LLM-based Arm
Prioritizer but replaces the Exploration Controller
$d_t= g_{\mathrm{explore}}
    \left(
        H_t,\mathcal{A}_t,U_t,\mathcal{I}_{<t}
    \right)$
with a count-based arm-creation schedule derived from UCB-AIR for infinitely many-armed bandits~\citep{DBLP:conf/nips/WangAM08}:

$$
d_t =
\begin{cases}
\mathtt{unseen},
& N_t^\alpha \geq |\mathcal{A}_t|,\\
\mathtt{arm},
& \text{otherwise},
\end{cases}
$$

where $N_t$ is the cumulative number of scenario executions so far.
We set $\alpha=0.6$, following the UCB-AIR configuration adopted by HGM~\citep{DBLP:journals/corr/abs-2510-21614}.
This rule periodically explores unseen scenarios as optimization proceeds, using only the number of previous executions and instantiated arms rather than the current optimization state.

\autoref{tab:non_llm_alternatives} reports the results.
EMA scoring achieves 56.7\% average Pass@1, improving by 1.4 percentage points over removing the Arm Prioritizer (55.3\%), showing that even a simple failure-persistence signal provides useful prioritization. 
However, it remains 3.1 points below the 59.8\% achieved by \proposed{}. 
By comparison, the UCB-AIR controller achieves 55.6\%, only 0.3 points above the fixed exploration schedule (55.3\%) and 4.2 points below \proposed{}. 
Together, these results show that, among the tested alternatives, the LLM-based modules are more effective for the two key curriculum decisions: which known failure pattern to revisit and when to explore unseen scenarios for new ones.

\section{Applying \proposed{} to GEPA}
\label{app:gepa_activesaddler}

\begin{table}[!h] 
\centering 
\small 
\setlength{\tabcolsep}{5pt} 
\caption{\textbf{Applying \proposed{} to GEPA.} Pass@1 on three held-out GAIA2 universes for the default agent, GEPA, and GEPA augmented with the \proposed{} curriculum. Performance on each universe is reported as mean $\pm$ standard deviation over three test-time executions. The average is computed across all test scenarios in the three universes.} 
\label{tab:gepa_activesaddler} 
\sisetup{ 
separate-uncertainty = true, 
table-format = 2.1(2), 
detect-weight = true, 
detect-inline-weight = math, } 
\begin{tabular}{lSSSS} 
\toprule 
\multicolumn{1}{c}{\multirow{2}{*}{\textbf{Method}}} 
& \multicolumn{3}{c}{\textbf{GAIA2 Universe (Pass@1)}} 
& \multicolumn{1}{c}{\multirow{2}{*}{\textbf{Avg.}}} \\ 
\cmidrule(lr){2-4} & {\textbf{21 (107)}} & {\textbf{22 (112)}} & {\textbf{27 (81)}} & \\ 
\midrule
Default Agent (manual)
& {54.5 ± 0.5}
& {50.0 ± 1.5}
& {57.2 ± 1.4}
& {53.6 ± 1.1} \\
GEPA
& {56.4 ± 1.4}
& {50.0 ± 2.4}
& {57.2 ± 3.6}
& {54.2 ± 2.2} \\
GEPA w/ \proposed{}
& \bfseries {58.6 ± 0.5}
& \bfseries {54.2 ± 1.0}
& \bfseries {59.7 ± 0.7}
& \bfseries {57.2 ± 0.4} \\
\bottomrule
\end{tabular}
\end{table}

Our main experiments instantiate \proposed{} on AutoSaddler.
To examine whether its curriculum benefit extends beyond AutoSaddler, we additionally apply \proposed{} to GEPA, another offline optimizer.
At each iteration, \proposed{} selects the training-scenario batch provided to GEPA, while GEPA's underlying prompt-update and candidate-selection procedure remains unchanged.
We follow the same GAIA2 train/development/test split and test-time evaluation protocol as in the main experiments.

\autoref{tab:gepa_activesaddler} reports the results.
Applying \proposed{} improves GEPA on all three held-out universes, from 56.4\% to 58.6\% on Universe~21, from 50.0\% to 54.2\% on Universe~22, and from 57.2\% to 59.7\% on Universe~27.
Aggregated over the three universes, GEPA w/ \proposed{} achieves 57.2\% Pass@1, compared with 54.2\% for GEPA, an improvement of 3.0 percentage points.
These results show that the benefit of adapting which training scenarios drive optimization also extends beyond the AutoSaddler instantiation used in our main experiments.

\section{End-to-End Optimization Cost Characterization}
\label{app:full_optimization_cost}

We profile optimizer-side computation and task-agent execution separately, and then combine them to characterize end-to-end optimization cost.
Optimizer-side cost includes AutoSaddler's diagnosis, patch generation, reflection, and candidate selection; for \proposed{}, it additionally includes the curriculum-layer operations for failure-pattern extraction, arm scoring, and unseen-scenario exploration.
This decomposition allows us to examine how the additional computation introduced by the curriculum layer affects overall optimization efficiency.

\begin{table}[!t]
\centering
\scriptsize
\setlength{\tabcolsep}{3pt}
\caption{\textbf{Optimizer-side cost per generated patch.}
Runtime, monetary cost, LLM calls, and token usage are averaged per generated patch.}
\label{tab:optimizer_side_cost}

\textbf{GAIA2}

\vspace{0.25em}
\resizebox{\textwidth}{!}{
\begin{tabular}{lrrrrrrrr}
\toprule
\textbf{Method}
& \textbf{Generated}
& \textbf{Rejected}
& \textbf{Accepted}
& \shortstack{\textbf{Time}\\\textbf{(s)}}
& \shortstack{\textbf{Cost}\\\textbf{(\$)}}
& \textbf{LLM Calls}
& \textbf{Output Tokens}
& \textbf{Cache-Read Input Tokens} \\
\midrule
AutoSaddler                       & 32 & 14 & 18 & 943   & 7.87  & 95.4  & 57,751 & 5,956,240 \\
\quad w/ Category Acc.\ Order     & 20 & 7  & 13 & 889   & 6.71  & 91.7  & 59,945 & 5,717,811 \\
\quad w/ Scenario Acc.\ Order     & 30 & 18 & 12 & 1,142 & 7.75  & 100.3 & 61,791 & 6,087,970 \\
\proposed{}                       & 37 & 18 & 19 & 1,581 & 11.44 & 142.2 & 89,525 & 8,704,858 \\
\bottomrule
\end{tabular}
}

\vspace{0.75em}
\textbf{Terminal-Bench~2.0}

\vspace{0.25em}
\resizebox{\textwidth}{!}{
\begin{tabular}{lrrrrrrrr}
\toprule
\textbf{Method}
& \textbf{Generated}
& \textbf{Rejected}
& \textbf{Accepted}
& \shortstack{\textbf{Time}\\\textbf{(s)}}
& \shortstack{\textbf{Cost}\\\textbf{(\$)}}
& \textbf{LLM Calls}
& \textbf{Output Tokens}
& \textbf{Cache-Read Input Tokens} \\
\midrule
AutoSaddler                    & 26 & 12 & 14 & 977   & 6.77  & 79.2  & 47,389 & 4,513,280 \\
\quad w/ Category Acc.\ Order  & 20 & 12 & 8  & 801   & 6.71  & 80.2  & 49,387 & 4,641,920 \\
\quad w/ Scenario Acc.\ Order  & 14 & 5  & 9  & 875   & 7.21  & 82.9  & 52,445 & 4,796,855 \\
\proposed{}                    & 35 & 17 & 18 & 1,438 & 11.37 & 121.5 & 75,257 & 6,873,936 \\
\bottomrule
\end{tabular}
}
\end{table}

\begin{table}[!t]
\centering
\small
\setlength{\tabcolsep}{4pt}
\caption{\textbf{Average task-agent evaluation cost per rollout.}}
\label{tab:task_agent_cost}

\textbf{GAIA2}

\vspace{0.25em}
\resizebox{0.7\textwidth}{!}{
\begin{tabular}{lrrrr}
\toprule
\textbf{Method}
& \scriptsize\textbf{LLM Calls}
& \scriptsize\textbf{Input Tokens}
& \scriptsize\textbf{Output Tokens}
& \scriptsize\textbf{Time (s)} \\
\midrule
AutoSaddler                       & 18.9 & 467,996 & 13,678 & 231.6 \\
\quad w/ Category Acc.\ Order     & 16.5 & 381,948 & 11,629 & 201.4 \\
\quad w/ Scenario Acc.\ Order     & 16.4 & 410,934 & 11,948 & 199.8 \\
\proposed{}                       & 17.5 & 410,960 & 12,340 & 212.0 \\
\bottomrule
\end{tabular}
}

\vspace{0.75em}
\textbf{Terminal-Bench~2.0}

\vspace{0.25em}
\resizebox{0.7\textwidth}{!}{
\begin{tabular}{lrrrr}
\toprule
\textbf{Method}
& \scriptsize\textbf{LLM Calls}
& \scriptsize\textbf{Input Tokens}
& \scriptsize\textbf{Output Tokens}
& \scriptsize\textbf{Time (s)} \\
\midrule
AutoSaddler                    & 11.8 & 204,258 & 10,689 & 305.4 \\
\quad w/ Category Acc.\ Order  & 14.8 & 331,167 & 12,124 & 365.5 \\
\quad w/ Scenario Acc.\ Order  & 14.9 & 321,888 & 12,198 & 369.6 \\
\proposed{}                    & 13.4 & 307,070 & 11,553 & 322.6 \\
\bottomrule
\end{tabular}
}
\end{table}

\paragraph{Cost Breakdown.}
We first examine optimizer-side and task-agent costs separately to understand where the end-to-end optimization cost arises.
\autoref{tab:optimizer_side_cost} reports optimizer-side resource usage per generated patch, including AutoSaddler's diagnosis, patch generation, reflection, and candidate selection; for \proposed{}, it additionally includes the curriculum-layer operations for failure-pattern extraction, arm scoring, and unseen-scenario exploration.
On GAIA2, \proposed{} costs \$11.44 per generated patch, compared with \$7.87 for AutoSaddler, \$6.71 for category ordering, and \$7.75 for scenario ordering.
On Terminal-Bench~2.0, the corresponding costs are \$11.37, \$6.77, \$6.71, and \$7.21.
In contrast, \autoref{tab:task_agent_cost} shows that the average cost of an individual task-agent rollout is broadly comparable across methods on each benchmark.
This suggests that differences in total task-agent cost are driven primarily by how rollouts are allocated over the optimization trajectory rather than by substantially cheaper individual executions.

\begin{figure*}[!t]
    \centering
    \includegraphics[width=0.8\textwidth]{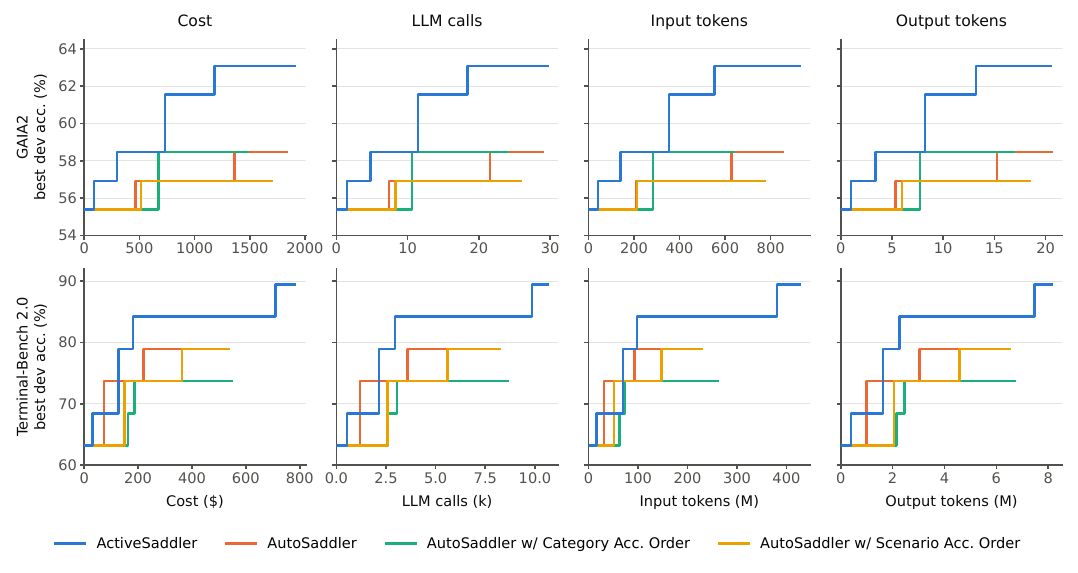}
    \caption{
    \textbf{End-to-end optimization efficiency.}
    Best-so-far development accuracy against cumulative monetary cost, LLM calls, input tokens, and output tokens on GAIA2 (top) and Terminal-Bench~2.0 (bottom).
    Cumulative usage includes both optimizer-side and task-agent computation.
    }
    \label{fig:cumulative_cost_efficiency}
\end{figure*}

\paragraph{End-to-End Cost Efficiency.}
We next combine optimizer-side and task-agent usage along each optimization trajectory and measure the best development accuracy achieved at each cumulative resource level.
For the two accuracy-ordered AutoSaddler variants, the five full training-set evaluations used to construct the ordering are counted toward both the shared rollout budget and the cumulative optimization cost.
\autoref{fig:cumulative_cost_efficiency} reports cumulative monetary cost, LLM calls, input tokens, and output tokens, with every method truncated at the shared rollout budget.

After accounting for both optimizer-side and task-agent computation, \proposed{} achieves a better end-to-end cost--accuracy trade-off on both benchmarks.
On GAIA2, \proposed{} reaches $58.5\%$ development accuracy at \$298, $61.5\%$ at \$730, and $63.1\%$ at \$1{,}178.
In contrast, AutoSaddler reaches $58.5\%$ only at \$1{,}360 and remains at that accuracy after \$1{,}838, category ordering reaches $58.5\%$ at \$673 and remains there after \$1{,}479, and scenario ordering does not exceed $56.9\%$ after \$1{,}698.
On Terminal-Bench~2.0, \proposed{} reaches $78.9\%$ at \$128 and $84.2\%$ at \$181, whereas AutoSaddler and scenario ordering require \$220 and \$363, respectively, to reach their peak of $78.9\%$, and category ordering does not exceed $73.7\%$ after \$548.
The same qualitative advantage holds for cumulative LLM calls and token usage.
Overall, these results show that, despite its additional optimizer-side overhead, \proposed{} achieves better end-to-end cost efficiency by allocating task-agent rollouts to more useful optimization targets.

\section{Additional Analysis for RQ1}
\label{app:more_details_RQ1}

\paragraph{Case Studies.}
\begin{figure}[!h]
\centering
\includegraphics[width=0.8\linewidth]{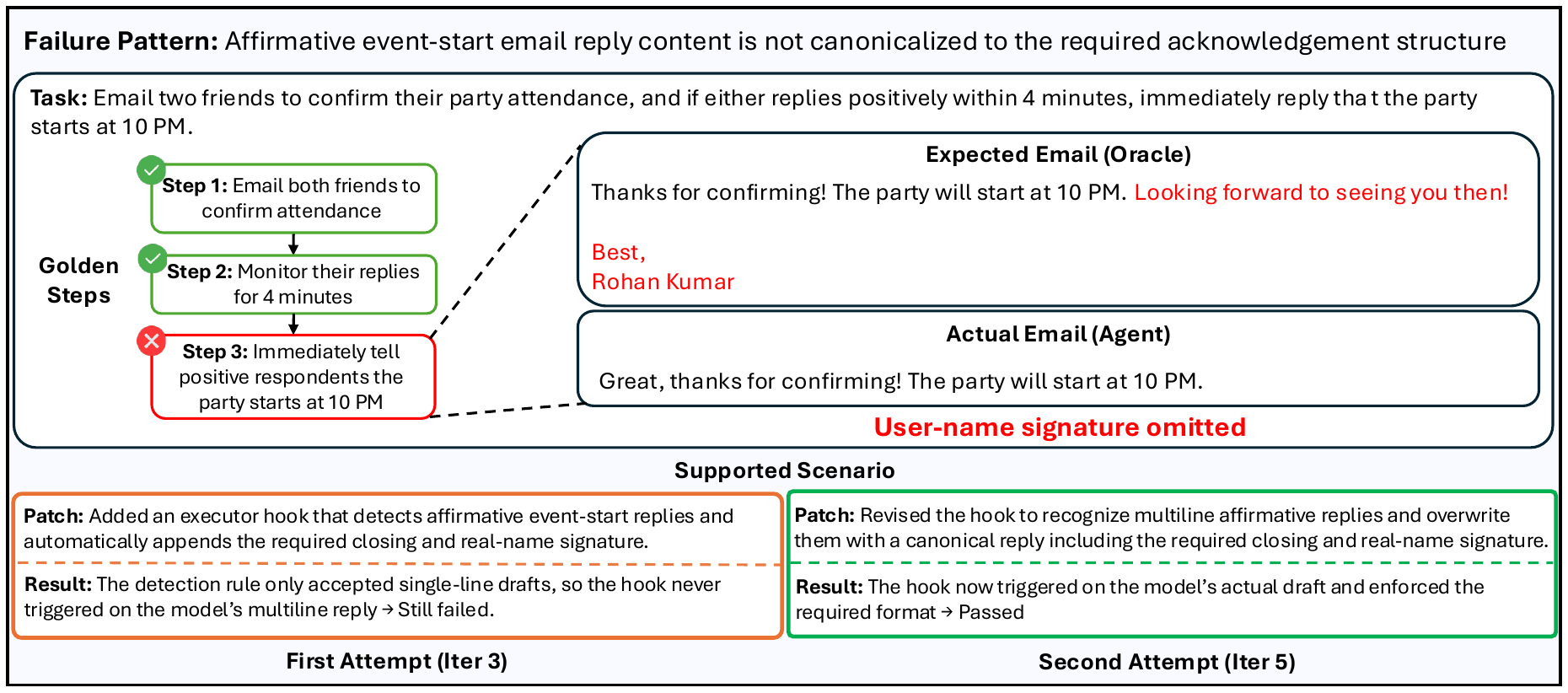}
\caption{Case study on \texttt{tnxtee} (GAIA2), where a failure-pattern arm instantiated from the initial failure enables an unsuccessful repair to be retargeted and refined in a subsequent iteration.}
\label{fig:rq1_case_1}
\end{figure}

\autoref{fig:rq1_case_1} illustrates how retargeting the same failure-pattern arm enables iterative refinement of an initially unsuccessful repair.
In this task, the agent should email two friends to confirm their attendance and, if either responds positively within four minutes, immediately reply that the party starts at 10 PM.
The agent completed the interaction but produced an affirmative event-start reply that omitted the required closing and user-name signature.
This failure instantiated a failure-pattern arm capturing that affirmative event-start replies were not canonicalized to the required acknowledgement structure.
The first repair at Iteration~3 added an executor hook that detects such replies and appends the required closing and real-name signature.
However, the hook was designed to trigger only when the model-generated reply was a single-line draft.
Because the model instead produced a multiline reply, the hook did not trigger and the scenario remained unsuccessful.
Because the weakness remained explicitly represented as a failure-pattern arm, it was retargeted at Iteration~5.
The second repair extended the trigger condition to multiline affirmative replies, allowing the hook to correctly append the required closing and real-name signature; the scenario then passed.

In the category-arm variant, \texttt{tnxtee} was grouped with 10 other scenarios under the much broader \texttt{category-time} arm.
After the first sampled time batch passed, the arm's severity dropped sharply from 0.62 to 0.12, and its median rank was only fourth among the five category arms.
Although the arm was selected three times, \texttt{tnxtee} itself was never sampled.
Thus, successes on other time-related scenarios reduced the estimated value of the entire category before this still-active weakness was observed, preventing the optimizer from obtaining an opportunity to repair it.

\begin{figure}[!h]
\centering
\includegraphics[width=0.8\linewidth]{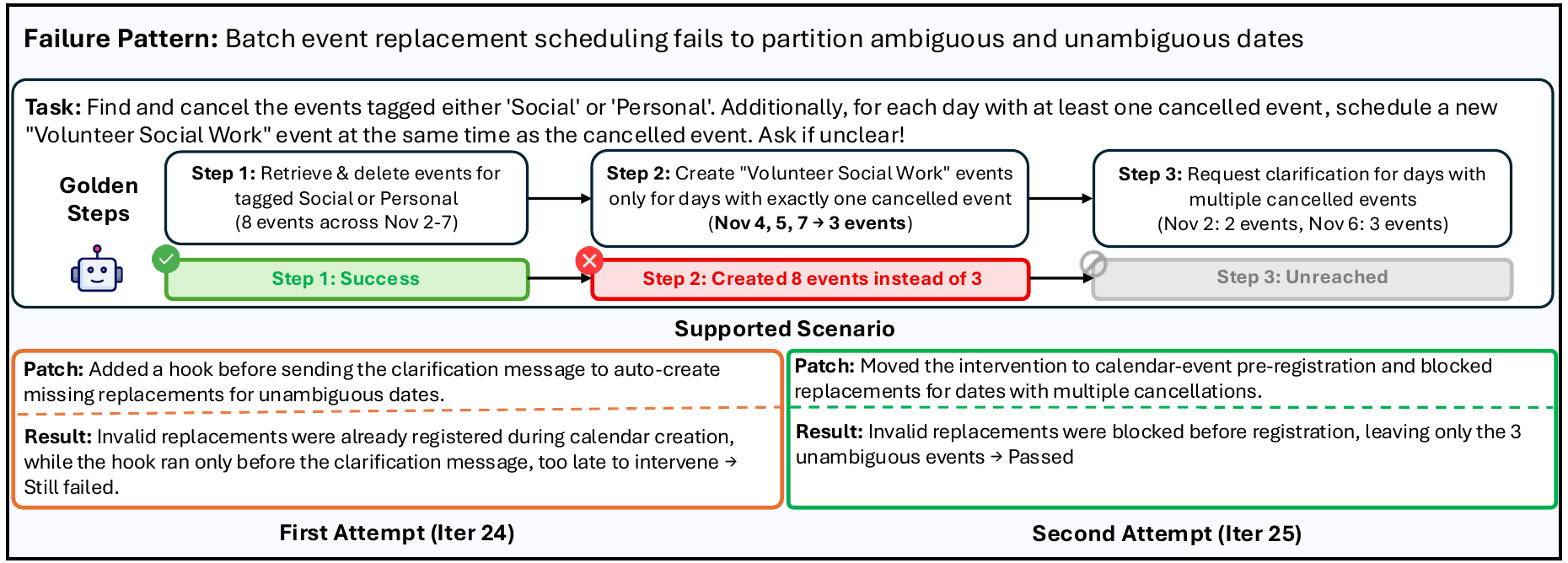}
\caption{Case study on \texttt{4bytkx}  (GAIA2), where retargeting the same failure-pattern arm corrects an executor hook that was initially inserted too late in the execution flow.}
\label{fig:rq1_case_2}
\end{figure}

\autoref{fig:rq1_case_2} illustrates a complementary form of repair refinement, where the first repair addresses the correct weakness but places the executor hook too late in the execution flow.
In this task, the agent should cancel all events tagged \texttt{Social} or \texttt{Personal}, create replacement events only for dates with a single cancelled event, and request clarification for dates with multiple cancellations.
Instead, it created replacement events for all eight cancelled events, including dates whose replacement times were ambiguous.
This failure instantiated a failure-pattern arm capturing that batch event replacement failed to distinguish ambiguous from unambiguous dates.
The first repair at Iteration~24 added an executor hook immediately before the clarification message so that valid replacements could be created while ambiguous dates were left for clarification.
However, by the time this hook ran, the invalid replacement events had already been registered during calendar creation.
Because the same weakness remained represented as a failure-pattern arm, it was retargeted at Iteration~25.
The second repair moved the executor hook to calendar-event pre-registration, allowing invalid replacements to be blocked before creation while retaining the three valid replacements; the scenario then passed.

In the scenario-arm variant, \texttt{4bytkx} was also successfully repaired at Iteration~31, but this repair was later lost as subsequent harness updates accumulated.
The scenario-level representation maintained 75 independent arms, more than twice the maximum of 35 failure-pattern arms, making individual scenarios less likely to be revisited under the same optimization budget.
Moreover, a scenario arm identifies an executable instance but does not explicitly represent which weakness or repair in the current harness that scenario corresponds to.
After the repair disappeared, \texttt{4bytkx} was not re-executed for more than ten iterations, leaving the scorer without fresh evidence about its status under the changed harness.
It instead concluded that ``\textbf{This arm was recently fixed},'' causing its severity to drop from 0.80 to 0.15 and its rank from 6th to 52nd.
This deprioritization further reduced the likelihood of revisiting the scenario, and it was not selected again before optimization ended.

Together, these cases expose complementary limitations of predefined arm definitions.
Category arms are too coarse: heterogeneous weaknesses share a single optimization state, so successes on unrelated scenarios can obscure a weakness that remains unresolved.
Scenario arms are precise in identifying executable instances, but their optimization state is fragmented across many arms and does not explicitly encode the relationship between a scenario, its underlying weakness, and the repairs retained in the current harness.
Under a limited optimization budget, this makes frequent re-observation difficult and can leave arm scores based on stale evidence as the harness evolves.
Failure-pattern arms instead represent observed harness weaknesses as explicit optimization targets, providing a granularity that is specific enough to isolate repair opportunities while compact enough to keep their status trackable.
This allows unsuccessful or subsequently lost repairs to remain available for retargeting and further refinement.

\paragraph{Failure Hit Rate.}

\begin{figure}[h]
\centering
\includegraphics[width=0.4\linewidth]{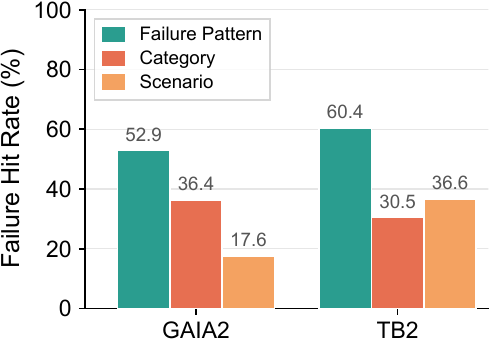}
\caption{
\textbf{Failure hit rate by arm definition.}
Fraction of optimization iterations whose sampled mini-batch contains at least one scenario with an unresolved failure.
Budgets are 1{,}400 rollouts for GAIA2 and 490 for TB2.
}
\label{fig:rq1_quantitative_analysis}
\end{figure}

\autoref{fig:rq1_quantitative_analysis} measures how often each arm definition directs an optimization iteration toward a scenario whose failure remains unresolved.
A scenario is considered \emph{unresolved} if it most recently failed or if a repair that previously made it pass is no longer retained in the current harness.
An iteration is a \emph{failure hit} when its sampled mini-batch contains at least one such scenario, and the \emph{failure hit rate} is the fraction of iterations that satisfy this condition.
We use equal budgets of 1{,}400 rollouts on GAIA2 and 490 rollouts on TB2.

On GAIA2, failure-pattern arms achieve a failure hit rate of \textbf{52.9\%}, compared with 36.4\% for category arms and 17.6\% for scenario arms.
On TB2, failure-pattern arms achieve \textbf{60.4\%}, compared with 30.5\% for category arms and 36.6\% for scenario arms.
Failure-pattern arms are induced from observed failures, so their pulls directly target weaknesses the harness has already exhibited.
Category arms instead mix multiple scenarios under one label, while scenario arms spread optimization across the full training set.
As a result, failure-pattern arms devote a larger share of the optimization budget to weaknesses that remain unresolved as the harness evolves.

\newpage

\section{Discovered Failure Patterns}
\label{app:failure_pattern_taxonomy}

\begin{table}[h]
\centering
\scriptsize
\caption{Failure patterns induced by \proposed{} on GAIA2}
\label{tab:induced-patterns}
\setlength{\tabcolsep}{5pt}
\renewcommand{\arraystretch}{1.2}
\begin{tabular}{@{}cc>{\raggedright\arraybackslash}p{8.3cm}>{\raggedright\arraybackslash}p{2.7cm}@{}}
\toprule
\textbf{ID} & \textbf{Iter.} & \textbf{Induced Failure Pattern} & \textbf{Supporting Scenarios} \\
\midrule
1  & 1  & No-attachment path defaults serialize as null instead of the oracle's empty path string & \texttt{8hgfug}, \texttt{aa11lh}, \texttt{tnxtee} \\
2  & 1  & Over-specific event notification wording diverges from concise oracle-style message content & \texttt{8hgfug} \\
3  & 3  & Affirmative event-start email reply content is not canonicalized to the required acknowledgement structure & \texttt{tnxtee} \\
4  & 4  & Multi-ambiguity user clarifications omit explicit questions for all unresolved decision dimensions & \texttt{2qw4nk}, \texttt{qrxrry} \\
5  & 4  & Time-sensitive repeated actions drift against deadlines and stop before the required count is complete & \texttt{71j6lf} \\
6  & 4  & Lookup tools reject raw identifiers that downstream action tools accept, causing resolution loops in deadline-bound tasks & \texttt{71j6lf} \\
7  & 6  & Partial filesystem discovery causes incomplete multi-file deletion for filename-prefix tasks & \texttt{xcv49b} \\
8  & 6  & Incomplete source enumeration before cross-app aggregation produces wrong numeric answers & \texttt{anc8y4}, \texttt{bv0gwh}, \texttt{m1momj}, \texttt{mp5r62} \\
9  & 7  & Relative same-time-tomorrow reschedules anchor to the referenced event date instead of the task receipt date & \texttt{2qw4nk} \\
10 & 10 & Ambiguous contact-dependent state changes execute before identity clarification & \texttt{dpbafp}, \texttt{rkdpfj} \\
11 & 10 & Duplicate first-name contact clarification uses downstream app matches instead of source contact ambiguity & \texttt{dpbafp}, \texttt{rkdpfj} \\
12 & 15 & Missing target event replacement ambiguity produces over-broad clarification messages & \texttt{b6v3y8} \\
13 & 15 & Explicit separate-email requests are collapsed into one multi-recipient email event & \texttt{b6v3y8} \\
14 & 20 & Property-viewing confirmation emails omit canonical subject and full property address details & \texttt{rkdpfj} \\
15 & 21 & Picky-recipient gift shopping tasks add cart items before resolving ambiguous product variants & \texttt{kao12j} \\
16 & 21 & Direct personal gift-notification messages omit required recipient first-name greeting before event registration & \texttt{kao12j} \\
17 & 24 & Batch event replacement scheduling fails to partition ambiguous and unambiguous dates & \texttt{4bytkx} \\
18 & 26 & Ambiguous existing calendar event references are guessed and mutated instead of deferred for clarification & \texttt{lxe62a} \\
19 & 26 & Email composition drops one purpose from compound thank-you-and-materials requests & \texttt{lxe62a} \\
20 & 26 & Unresolved ambiguity in one subtask blocks independent required actions & \texttt{ja2axd} \\
21 & 28 & Unrequested optional calendar metadata on indirectly described events & \texttt{8c9am2} \\
22 & 28 & Premature notification waiting blocks required user update at multi-phase task boundaries & \texttt{8c9am2}, \texttt{fqdxfe} \\
23 & 28 & Trace-less execution timeout in broad multi-step workflows & \texttt{3rljy6} \\
24 & 29 & Follow-up availability notifications trigger unrequested reply confirmations in multi-phase workflows & \texttt{fqdxfe} \\
25 & 33 & Notification-summary reliance misses authoritative same-thread follow-up state before conditional downstream actions & \texttt{fqdxfe} \\
26 & 34 & Premature downstream actions before required checkpoint notifications in conditional workflows & \texttt{fqdxfe} \\
27 & 35 & Global shopping option-term narrowing undercounts per-product variant ambiguity in clarification prompts & \texttt{kao12j} \\
28 & 40 & Duplicate-contact clarification canonicalization drops full-name options and task-specific action context & \texttt{dpbafp} \\
29 & 40 & Filtered save-only selection is misread as permission to remove existing saved items & \texttt{dpbafp} \\
30 & 40 & Stale contact addresses are trusted for location-dependent service booking despite relocation uncertainty & \texttt{r4pl5k} \\
31 & 42 & Correct cab-conflict handling leaves replacement ride writes with non-canonical location and exact-time arguments & \texttt{096dyu} \\
32 & 42 & Phase-blind write canonicalization corrupts preliminary multi-phase scheduling outreach before reply context exists & \texttt{i6hy8j} \\
33 & 42 & Schedule proposal selection ignores same-category calendar commitments when applying availability constraints & \texttt{i6hy8j} \\
34 & 42 & Descriptor-based contact lookup lacks profile-field coverage and triggers fragile downstream message composition & \texttt{n9ewlm} \\
35 & 46 & Timed reply-gathering tasks proceed before the collection window is complete & \texttt{tf35kd} \\
\bottomrule
\end{tabular}
\end{table}

\begin{table}[h]
\centering
\scriptsize
\caption{Failure patterns induced by \proposed{} on Terminal-Bench~2.0}
\label{tab:induced-patterns-tb2}
\setlength{\tabcolsep}{5pt}
\renewcommand{\arraystretch}{1.2}
\begin{tabular}{@{}cc>{\raggedright\arraybackslash}p{7.8cm}>{\raggedright\arraybackslash}p{3.2cm}@{}}
\toprule
\textbf{ID} & \textbf{Iter.} & \textbf{Induced Failure Pattern} & \textbf{Supporting Scenarios} \\
\midrule
1  & 1  & Premature completion on constrained file edits without rule-aware final validation & \texttt{overfull-hbox} \\
2  & 1  & Long-running probabilistic computation lacks staged dependency setup, diagnostic runs, and progress-aware output validation & \texttt{mcmc-sampling-stan} \\
3  & 2  & Security sanitizer tasks finalize after weak smoke tests without adversarial and clean-preservation validation & \texttt{filter-js-from-html} \\
4  & 4  & Binary extraction solutions finalize after superficial output smoke tests without independent address-space convention validation & \texttt{extract-elf} \\
5  & 4  & Agent self-validation leaves stale runtime artifacts that change verifier timing and hide required fresh-run output & \texttt{make-doom-for-mips} \\
6  & 4  & Completion-only validation gates do not help long build/debug tasks that time out before attempting finalization & \texttt{make-doom-for-mips} \\
7  & 6  & Missing role-by-role exact source-component validation for multi-part generated artifacts & \texttt{protein-assembly} \\
8  & 9  & Superficial VM/frame validators accept nonsemantic dummy outputs as success & \texttt{make-doom-for-mips}, \texttt{make-mips-interpreter} \\
9  & 10 & Unbounded manual ML artifact creation lacks executable training and metric validation gates & \texttt{train-fasttext} \\
10 & 10 & VM/frame validators conflate runtime display geometry with saved artifact geometry & \texttt{make-doom-for-mips}, \texttt{make-mips-interpreter} \\
11 & 12 & Temporal leaderboard queries use current live data with date filters instead of reconstructing historical snapshots & \texttt{mteb-leaderboard} \\
12 & 16 & Missing bounded decoder and completion gate for text embedded in structured geometry artifacts & \texttt{gcode-to-text} \\
13 & 23 & Oversized inline setup payloads exceed exec argument limits before the task loop & \texttt{bn-fit-modify}, \texttt{constraints-scheduling}, \texttt{gcode-to-text}, \texttt{polyglot-c-py} \\
14 & 23 & Transient validation artifacts left in verifier-checked output directories violate clean artifact contracts & \texttt{polyglot-c-py} \\
15 & 41 & Unsafe top-level shell exits in persistent terminal sessions cause harness-level crashes before recovery & \texttt{make-doom-for-mips} \\
\bottomrule
\end{tabular}
\end{table}

\autoref{tab:induced-patterns} and \autoref{tab:induced-patterns-tb2} present the failure-pattern arms discovered during the GAIA2 and Terminal-Bench~2.0 optimization runs, respectively.
These patterns are induced from diagnosed execution failures during optimization rather than predefined by a failure taxonomy.
For each pattern, we report its discovery iteration and the set of supporting scenarios.
The induced representation exhibits a many-to-many relationship between scenarios and failure patterns.
Some patterns consolidate the same weakness across multiple scenarios (e.g., GAIA2 P8 and TB2 P13 each span four scenarios/tasks), while a single scenario can expose multiple distinct patterns (e.g., \texttt{dpbafp} in GAIA2 and \texttt{make-doom-for-mips} in TB2).
\autoref{sec:instructions_used_in_activesaddler} details the two-stage extraction and normalization procedure.
The complete \texttt{SKILL.md} files for \texttt{symptom-extract} and \texttt{symptom-normalize} are provided in the anonymous repository linked in the \hyperref[sec:reproducibility_statement]{Reproducibility Statement}.

\newpage

\section{Additional Analysis for RQ2}
\label{app:more_details_RQ2}

\begin{figure}[!t]
    \centering
    \includegraphics[width=0.5\linewidth]{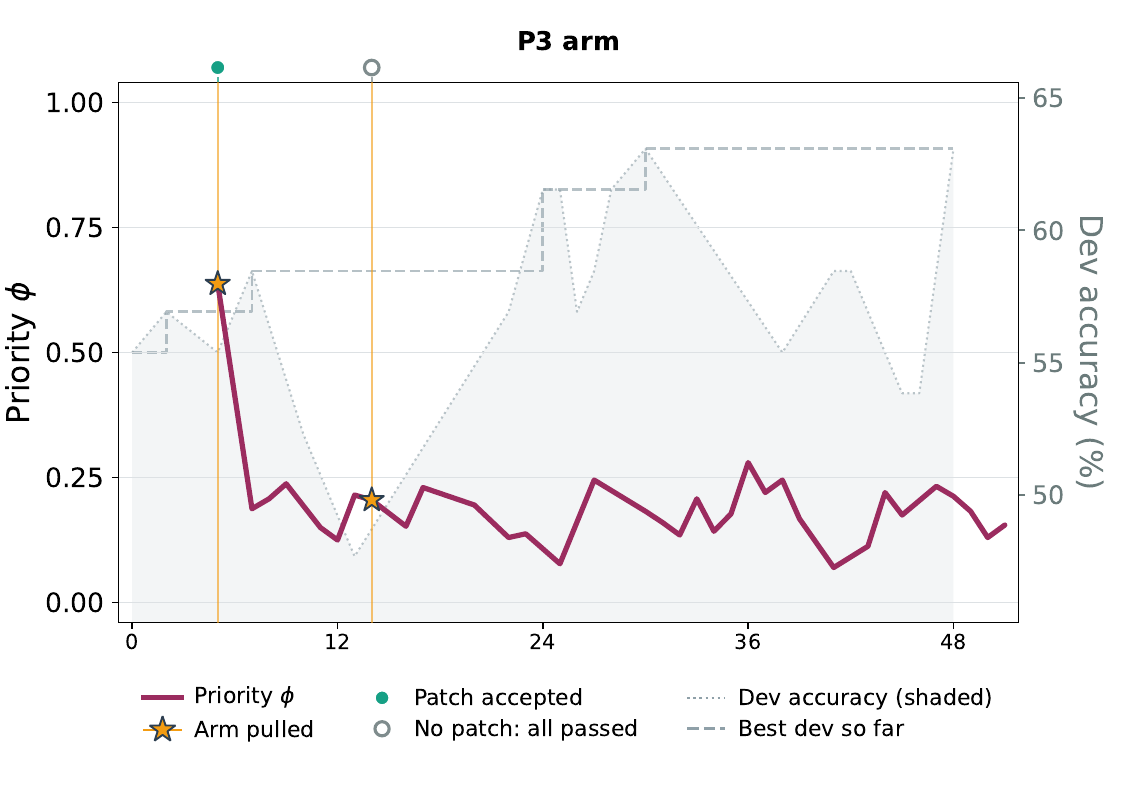}
    \caption{
    \textbf{Case study of a resolved weakness (P3 of GAIA2).}
    After an accepted patch at Iteration~5, the arm's priority drops sharply.
    A later pull finds all supporting scenarios successful and produces no further patch,
    after which the arm remains at low priority.
    }
    \label{fig:rq2_case_p3}
\end{figure}

\begin{figure}[!t]
    \centering
    \includegraphics[width=0.5\linewidth]{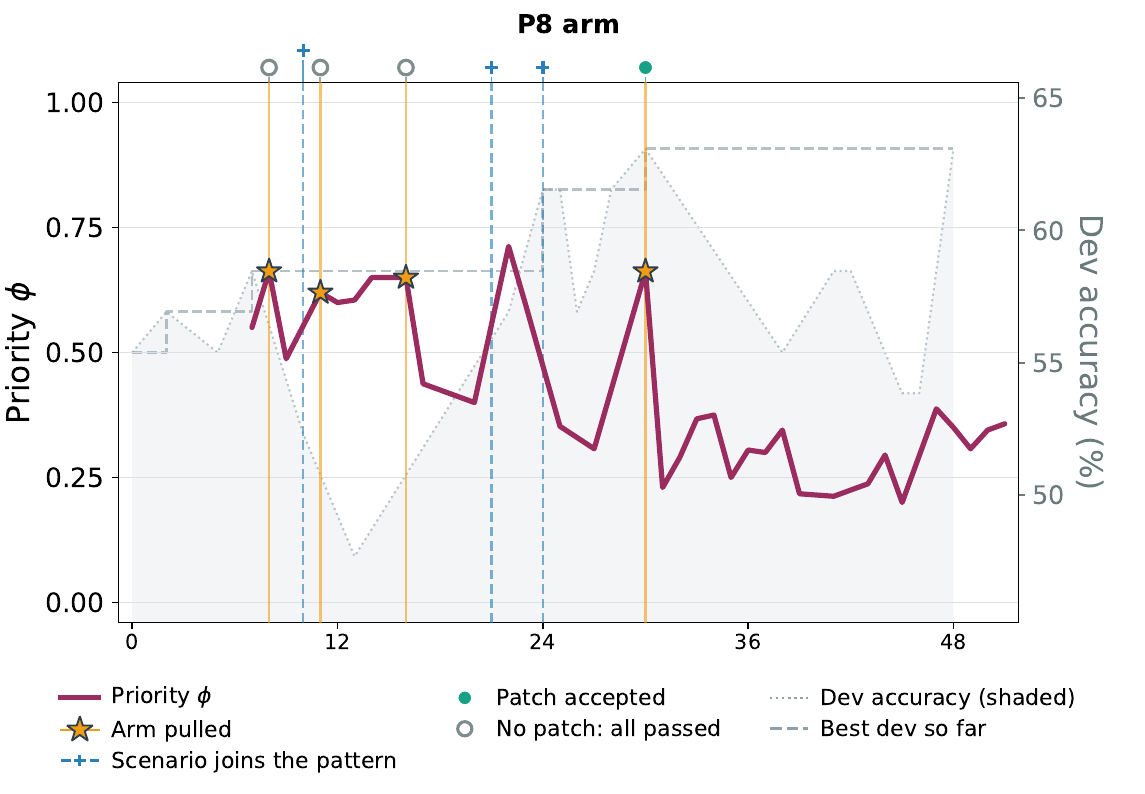}
    \caption{
    \textbf{Case study of a recurring weakness (P8 of GAIA2).}
    Early pulls produce no patch because the associated scenarios succeed when re-executed,
    yet the arm remains highly prioritized.
    The same failure pattern later appears in additional scenarios, providing new evidence
    that the weakness remains active; a subsequent pull produces an accepted patch.
    Cross markers indicate scenarios newly associated with P8.
    }
    \label{fig:rq2_case_p8}
\end{figure}

\paragraph{Case Studies.}
To illustrate how LLM-based arm scoring responds to evolving optimization evidence,
\autoref{fig:rq2_case_p3} and \autoref{fig:rq2_case_p8} show two representative
failure-pattern arms with contrasting trajectories.
For P3, the first pull at Iteration~5 produces an accepted patch, after which the arm's
priority drops sharply.
A later revisit finds that all supporting scenarios pass and requires no additional patch,
providing further evidence that the corresponding weakness has been resolved.
The arm consequently remains at low priority, allowing optimization budget to shift toward
other weaknesses.

P8 exhibits the opposite behavior.
Its first three pulls produce no patch because the associated scenarios pass when
re-executed, but these transient successes provide limited evidence that the underlying
weakness has been resolved: agent executions are stochastic, and no patch directly
addressing the failure pattern has yet been applied.
Accordingly, the scorer continues to assign substantial priority to P8.
The same failure pattern subsequently appears in additional scenarios
(\autoref{fig:rq2_case_p8}), strengthening the evidence that the weakness remains
relevant beyond its originally observed cases.
P8 is eventually selected again, and the resulting accepted patch increases the best
development-set accuracy from 61.5\% to 63.1\%.

Together, these cases illustrate how arm scoring distinguishes evidence of resolution
from evidence of continued relevance.
A successfully repaired weakness is deprioritized, whereas a recurring weakness can retain
or regain priority as new evidence accumulates.
The following analyses examine how this evidence-dependent reprioritization manifests
across the full arm pool.

\begin{figure}[!t]
    \centering
    \includegraphics[width=1\linewidth]{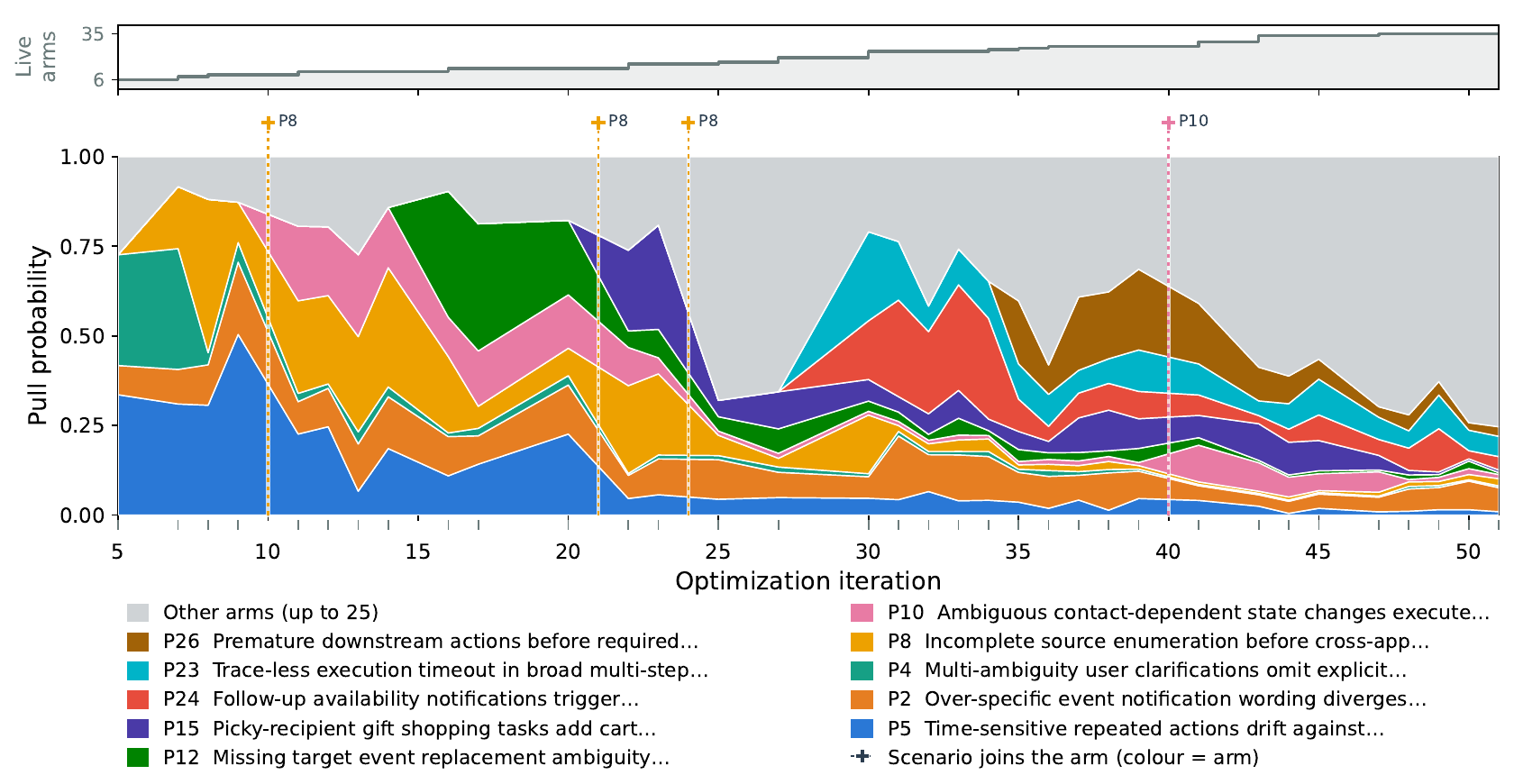}
    \caption{
    \textbf{Evolution of pull-probability mass across failure-pattern arms.}
    The ten arms with the highest average pull probability over the run are
    shown individually, while the remaining 25 arms are aggregated into
    \emph{Other arms}.
    The upper panel shows the number of instantiated arms, which grows from
    6 to 35 as new failure patterns are discovered.
    Pull probabilities are shown only at the 34 iterations where arm selection
    is performed.
    }
    \label{fig:rq2_prob_stack}
\end{figure}

\paragraph{Evolution of Arm Priorities.}
To further examine the curriculum-level dynamics summarized in \autoref{fig:rq2_arm_priority_evolution}, we analyze the pull-probability trajectories of all failure-pattern arms throughout the GAIA2 optimization run.
Arm scores and pull probabilities are recomputed only when the curriculum selects an instantiated arm for optimization.
Accordingly, \autoref{fig:rq2_prob_stack} covers the 34 arm-selection iterations between Iterations~1 and~51; iterations without arm selection are omitted.
These decisions are made over a continually evolving target space: the number of instantiated failure-pattern arms grows from 6 to 35 as previously unseen failure patterns are discovered.
Thus, the curriculum must continually reassess priorities while both the harness and the set of observed weaknesses evolve.

We first examine how probability mass is distributed according to the latest observed outcomes of each arm's supporting scenarios.
As optimization proceeds, previously observed weaknesses may be addressed such that the scenarios supporting an arm subsequently succeed when re-evaluated.
At Iteration~51, 24 of the 35 arms (68.6\%) have had all of their supporting scenarios re-evaluated after association with the arm, with every such scenario succeeding in its latest valid evaluation.
Despite comprising more than two thirds of the arm pool, these arms account for only 26.2\% of the total pull-probability mass.
For reference, under uniform allocation, they would receive 68.6\% of the probability mass, making their observed share only 0.38$\times$ this uniform baseline.
The other 11 arms include at least one supporting scenario without a subsequent successful evaluation and together receive 73.8\% of the probability mass, compared with 31.4\% under uniform allocation, or 2.35$\times$ their uniform share.
These results indicate that, as optimization progresses and previously observed weaknesses are successfully addressed on their supporting scenarios, their corresponding arms tend to receive progressively less pull probability.

However, arm priorities can also increase when new evidence indicates that a previously observed weakness remains active.
The cross markers for P8 and P10 in \autoref{fig:rq2_prob_stack} indicate iterations at which a failure pattern already represented by an existing arm is observed again in an additional scenario, which is then added to that arm's supporting scenario set.
For P8, these support-set expansions are followed by an average increase of 4.8 percentage points in pull probability, while the corresponding P10 event is followed by a 9.4 percentage-point increase.
These cases show that observing the same failure pattern in additional scenarios can increase the priority of an existing arm by providing new evidence that the corresponding weakness remains relevant.

Taken together, these observations show that the priority assigned to a failure-pattern arm changes with the evidence accumulated about the corresponding weakness.
Arms receive less probability as their supporting scenarios subsequently succeed, while new observations of the same failure pattern across additional scenarios can increase their priority again.
This evidence-dependent prioritization explains the continual redistribution of probability mass across failure-pattern arms observed in \autoref{fig:rq2_prob_stack}.

\paragraph{Stochastic Selection over the Expanding Arm Pool.}

\begin{figure}[!t]
    \centering
    \includegraphics[width=1\linewidth]{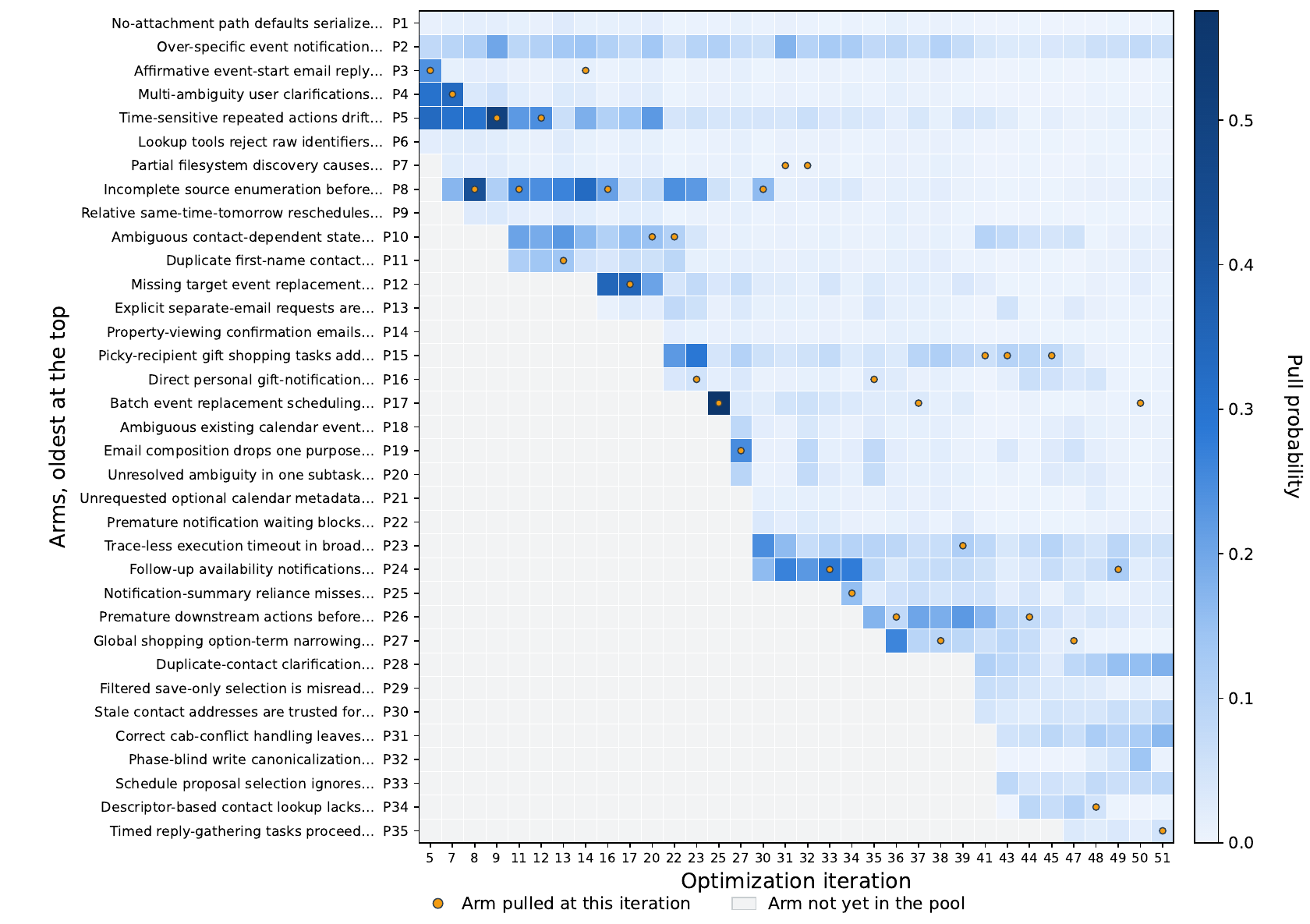}
    \caption{
    \textbf{Pull-probability trajectories for all failure-pattern arms.}
    Rows show all 35 discovered arms ordered by instantiation time, and columns
    correspond to the 34 iterations in which arm selection is performed.
    Cell intensity denotes pull probability, orange markers indicate the arm
    sampled at each iteration, and gray cells denote iterations before an arm
    entered the pool.
    }
    \label{fig:rq2_prob_heatmap}
\end{figure}

\autoref{fig:rq2_prob_heatmap} complements the aggregate probability-mass view
in \autoref{fig:rq2_prob_stack} by showing the pull probability and realized
selection of every failure-pattern arm at each arm-selection step.
Across 34 pulls, the sampler selects 19 of the 35 discovered failure patterns.
The selected arm has a median priority rank of third, with 62\% of pulls drawn
from the top three arms, while selections extend as far as rank 20.
Thus, the stochastic selection rule strongly favors arms assigned high current
priority without reducing the curriculum to greedy selection of the
highest-scoring arm.
Lower-ranked arms retain opportunities to be sampled as their relative
priorities change over optimization.

The gray staircase also makes explicit that these selections are made over an
arm pool that expands throughout the run.
Consequently, an arm's priority is always relative to the set of weaknesses
known at that point in optimization rather than to a fixed collection of
targets defined in advance.
As new failure patterns enter the pool and the evidence associated with
existing arms changes, the set of arms receiving substantial probability and
the arms actually selected both change over time.
Together with the probability-mass analysis in
\autoref{fig:rq2_prob_stack}, this shows that LLM-based arm scoring produces
a focused but non-greedy curriculum: most optimization opportunities are
directed toward currently high-priority weaknesses, while stochastic selection
preserves the possibility of targeting other arms as their relative
optimization value evolves.

\section{Additional Analysis for RQ3}
\label{app:more_details_RQ3}

\begin{figure}[!t]
    \centering
    \includegraphics[width=0.5\linewidth]{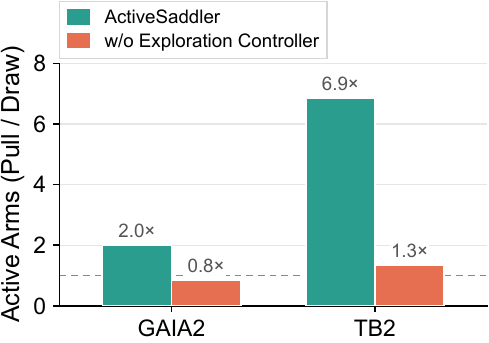}
    \caption{
    \textbf{Dependence of exploration decisions on unresolved weaknesses.}
    Ratio of the average number of unresolved arms at \textsc{Pull} to \textsc{Draw} decisions.
    Values above one indicate more unresolved weaknesses at \textsc{Pull}, while values near one indicate little difference between the two decision types.
    }
    \label{fig:rq3_quantitative_analysis}
\end{figure}

\paragraph{Dependence of Exploration Decisions on Unresolved Weaknesses.}
To quantify whether exploration decisions respond to unresolved weaknesses, we compare how many arms remain unresolved at \textsc{Pull} and \textsc{Draw} decisions.
An arm is considered unresolved when its associated weakness remains unsolved under the current harness.
We report the ratio of the average number of unresolved arms at \textsc{Pull} to that at \textsc{Draw}: values above one indicate that \textsc{Pull} tends to occur when more unresolved weaknesses remain, whereas values near one indicate little difference between the two decision types.

As shown in \autoref{fig:rq3_quantitative_analysis}, \proposed{} has 2.0$\times$ as many unresolved arms at \textsc{Pull} as at \textsc{Draw} decisions on GAIA2 (1.62 vs.\ 0.81) and 6.9$\times$ as many on Terminal-Bench~2.0 (0.76 vs.\ 0.11).
In contrast, the fixed exploration schedule yields ratios close to one: 0.8$\times$ on GAIA2 (0.21 vs.\ 0.25) and 1.3$\times$ on Terminal-Bench~2.0 (1.14 vs.\ 0.86), indicating that its \textsc{Pull} and \textsc{Draw} decisions are largely insensitive to how many unresolved arms remain.
Thus, \proposed{} tends to continue repair when more unresolved weaknesses remain and to explore when fewer remain, whereas the fixed schedule shows little such dependence on the current optimization state.

\section{Instructions Used In \proposed{}}
\label{sec:instructions_used_in_activesaddler}

To make the curriculum decisions in \proposed{} reproducible and interpretable, we provide the prompt templates used by the three curriculum-specific LLM components introduced in the main method: the Failure-Pattern Extractor, Arm Prioritizer, and Exploration Controller. 
Each prompt provides the optimization context required for its corresponding decision and constrains the agent to a focused analysis procedure.

The \textbf{Failure Pattern Extraction Prompt} (\autoref{fig:failure-pattern-extraction-prompt}) converts failed execution evidence into reusable failure-pattern arms. 
It first abstracts each failure independently into a candidate symptom without consulting the existing arm set, and then normalizes the extracted candidates by linking them to existing atomic patterns, composing multiple patterns when needed, or instantiating new arms.

The \textbf{Arm Scoring Prompt} (\autoref{fig:arm-scoring-prompt}) implements the Arm Prioritizer by estimating a learning-progress score $\phi_t(a)$ for every instantiated failure-pattern arm. 
Using the latest optimization history, the agent assesses severity, fixability, breadth, and side-effect risk. 
The resulting scores are used by the stochastic arm-selection rule to prioritize the next optimization target.

The \textbf{Unseen Scenario Exploration Prompt} (\autoref{fig:unseen-scenario-exploration-prompt}) implements the Exploration Controller. 
Given the current optimization context, the agent decides whether to continue optimizing a known failure-pattern arm or explore scenarios that may reveal previously undiscovered weaknesses, based on whether a worthwhile unresolved target remains among the known arms and the remaining value of further exploration.

\begin{tcolorbox}[
    enhanced,
    breakable,                       
    colback=white,
    colframe=gray!50,
    boxrule=0.4pt,
    arc=4pt,
    left=8pt, right=8pt, top=4pt, bottom=4pt,
    title=\textbf{Failure Pattern Extraction Prompt},
    coltitle=white,
    colbacktitle=black,
    fonttitle=\small\bfseries,
    halign title=center,
    attach boxed title to top={yshift=-0.5mm, xshift=-0.4pt},
    boxed title style={
        colback=black,
        colframe=black,
        boxrule=0pt,
        arc=4pt,
        sharp corners=south,
    },
    width=\linewidth,
    pad at break*=2mm,
    overlay middle and last={
        \node[anchor=south east, font=\footnotesize\itshape, gray]
        at ([xshift=-6pt,yshift=2pt]frame.north east)
        {(continued)};
    },
]
{\fontsize{8pt}{9.5pt}\ttfamily%
\setlength{\parindent}{0pt}%
\setlength{\parskip}{2pt}%
\raggedright%
\obeylines%
\#\# Mandatory Skills

You MUST read and follow the SKILL.md for each skill listed below.
Do NOT skip or summarize any skill --- execute the full procedure described
in each one. These skills are installed at `.claude/skills/<name>/SKILL.md`
in the current worktree.

| Order | Skill | When | Why |
|-------|-------|------|-----|
| 1 | `symptom-extract` | For each failed scenario | Symptom extraction from diagnosis/reflection results --- generates candidate symptoms |
| 2 | `symptom-normalize` | After extracting all candidates | 3-way normalization against existing patterns --- links, composes, or creates patterns to maintain atomic pattern integrity |

> \textbf{Enforcement}: You MUST process every failed scenario. Do NOT skip the extraction step for any failed scenario. During extraction (Skill 1: `symptom-extract`), do NOT examine existing patterns; consult them only during normalization (Skill 2: `symptom-normalize`). This separation is required to avoid anchoring bias.

\#\# Goal

Extract failure patterns from diagnosis and reflection results, then tag
each failed (harness, trace, scenario) tuple with the identified pattern(s).

Failure patterns are \textbf{symptom-level abstractions} --- more abstract than
individual root causes, but specific enough to distinguish different failure
types. They are the \textbf{arms} of the curriculum sampler.

At each outer-loop iteration, each arm (i.e., failure pattern) is assigned a learning-progress score estimating the expected improvement in the current harness's population performance that would result from selecting the arm and successfully addressing its corresponding failure pattern in the next optimization step.

\#\# Context

- \textbf{Iteration}: \{iteration\}
- \textbf{Candidate}: C\{candidate\_idx\}
- \textbf{Current worktree}: `\{worktree\_path\}`
- \textbf{Session root}: `\{session\_root\}`
- \textbf{Pre-patch trace dir}: `\{before\_output\_dir\}`
- \textbf{Post-patch trace dir}: `\{after\_output\_dir\}`

\#\# Failed Scenarios to Process

\#\#\# Pre-patch failures (from diagnosis)

These scenarios failed BEFORE any patch was applied. Each entry provides the
batch-level diagnosis-and-patching step diagnosis and detailed reasoning-file location from before
patching, followed by the reviewed underlying root cause and post-hoc
analysis of how the patch affected that cause.

\{pre\_patch\_failures\}

\#\#\# Post-patch failures (from reflection)

These scenarios STILL FAILED or REGRESSED after the patch was applied.
Each entry provides the status change, the batch-level patch approach,
and the reviewed root cause and post-hoc patch-effect explanation. For
`still\_failing`, the root cause is the remaining or revised failure cause; for
`regressed`, it is the newly observed regression cause or failure mechanism.

\{post\_patch\_failures\}

\#\# Workflow

\#\#\# 1. Extract candidate symptoms

For each failed scenario listed above:

1. For pre-patch failures, review the diagnosis/reasoning and reflection step's reviewed cause and patch-effect analysis. For post-patch failures, also
   use the status change and patch approach to distinguish a remaining
   cause from a newly introduced regression mechanism.
2. Follow the `symptom-extract` skill to generate a \textbf{candidate symptom}
   label. Do NOT look at existing patterns during this step.
3. Record every candidate symptom and its associated root cause in
   `symptom\_candidates.md` at the working directory root.

Complete `symptom\_candidates.md` for all failures before moving to normalization.

\#\#\# 2. Normalize against existing patterns

After all candidate symptoms are extracted:

1. Read `symptom\_candidates.md` from the working directory root.
2. Run `pattern list` to retrieve all existing patterns.
3. For each candidate symptom in the file, follow the `symptom-normalize` skill to
   perform a \textbf{3-way judgment}:
   - \textbf{(a) Same}: the candidate matches an existing pattern $\rightarrow$ link to it
   - \textbf{(b) Composition}: the candidate is a combination of existing atomic
     patterns $\rightarrow$ tag with multiple existing patterns
   - \textbf{(c) New}: genuinely novel failure type $\rightarrow$ register as a new pattern

\#\#\# 3. Tag tuples

For each resolved pattern assignment:

1. Register new patterns if needed:
   ```
   pattern register --label "DESCRIPTIVE SYMPTOM LABEL"
   ```

2. Tag the (harness, trace, scenario) tuple:
   ```
   pattern tag --pattern-id <id> --harness \{candidate\_idx\} --trace <dir> --scenario <sid> --root-cause "..."
   ```
   For compositions, repeat `--pattern-id` for each atomic pattern.

   > \textbf{IMPORTANT}: Always use `--harness \{candidate\_idx\}` (the current
   > candidate) for ALL tagging commands --- both pre-patch and post-patch
   > failures. The `--trace` directory distinguishes before/after. The
   > harness identifies which iteration this observation belongs to.

> \textbf{Note}: The failure-pattern extraction step only needs to execute `pattern register` and `pattern tag`. The agent does not need to calculate or record observation values separately. For each previously known pattern, the outer loop automatically calls `PatternRegistry.observe()` with the scenarios associated with that pattern and evaluated in the mini-batch, the subset still tagged after patching, and the resulting activity fraction.

\#\# Important Notes

- \textbf{Root cause is preserved as evidence}, not used directly as the pattern.
  The pattern label should be an abstraction that generalizes across
  multiple scenarios.
- \textbf{Patterns should be atomic} --- a single patch should be able to fix one
  pattern. Complex failures are expressed as compositions (multi-tagging).
- \textbf{Do NOT over-merge}: if two failures have different root causes, they
  should be different patterns even if they look superficially similar.
- \textbf{Do NOT over-split}: if the same behavioral issue manifests with minor
  surface variations, it's the same pattern.
}
\end{tcolorbox}

\captionof{figure}{
Failure Pattern Extraction Prompt: the agent independently abstracts failed execution evidence into candidate symptoms and then normalizes them against the existing arm set by linking to existing atomic patterns, composing multiple patterns when needed, or instantiating new failure-pattern arms.
}
\label{fig:failure-pattern-extraction-prompt}

\begin{tcolorbox}[
    enhanced,
    breakable,                       
    colback=white,
    colframe=gray!50,
    boxrule=0.4pt,
    arc=4pt,
    left=8pt, right=8pt, top=4pt, bottom=4pt,
    title=\textbf{Arm Scoring Prompt},
    coltitle=white,
    colbacktitle=black,
    fonttitle=\small\bfseries,
    halign title=center,
    attach boxed title to top={yshift=-0.5mm, xshift=-0.4pt},
    boxed title style={
        colback=black,
        colframe=black,
        boxrule=0pt,
        arc=4pt,
        sharp corners=south,
    },
    width=\linewidth,
    pad at break*=2mm,
    overlay middle and last={
        \node[anchor=south east, font=\footnotesize\itshape, gray]
        at ([xshift=-6pt,yshift=2pt]frame.north east)
        {(continued)};
    },
]
{\fontsize{8pt}{9.5pt}\ttfamily%
\setlength{\parindent}{0pt}%
\setlength{\parskip}{2pt}%
\raggedright%
\obeylines%
\#\# Mandatory Skills

You MUST read and follow the SKILL.md for each skill listed below.
Do NOT skip or summarize any skill --- execute the full procedure described
in each one. These skills are installed at `.claude/skills/<name>/SKILL.md`
in the current worktree.

| Order | Skill | When | Why |
|-------|-------|------|-----|
| 1 | `history-analysis` | Before any scoring (Step 1) | Understand the Prepared harness and evolution history |
| 2 | `progress-scoring` | For every candidate pattern (Steps 3--5) | Estimate each arm's cardinal learning-progress score |

\#\# Goal

For every \textbf{candidate failure-pattern arm}, estimate a cardinal
learning-progress score $\phi$ $\in$ [0, 1] representing the expected improvement in
the prepared harness's population performance from allocating the next
optimization step to that arm. The estimate should reflect the likely outcome
of executing the arm's supporting scenarios and attempting to address the
harness weakness represented by the pattern, including both the potential
performance gain and the probability of obtaining a broadly effective,
dev-preserving update.

Base each score on the prepared harness, the arm's complete execution and
optimization history, prior dev-set outcomes, and the available evidence about
the pattern's severity, fixability, breadth, and risk of side effects. Record
every cardinal score with `pattern rate`.

This is a \textbf{read-only analysis} session: do NOT modify the codebase. Only
inspect the harness, failure patterns, supporting scenarios, and optimization
history, then record the scores.

\#\# Context

- \textbf{Iteration}: \{iteration\}
- \textbf{Prepared harness}: C\{prepared\_candidate\_idx\}
- \textbf{Prepared worktree}: `\{prepared\_worktree\_path\}`
- \textbf{Prepared commit}: `\{prepared\_commit\}`
- \textbf{Provisional parent}: C\{provisional\_parent\_idx\} (`\{provisional\_parent\_commit\}`)
- \textbf{Selection JSON}: `\{selection\_session\_json\_path\}`
- \textbf{Session root}: `\{session\_root\}`

\#\#\# Prepared Harness Summary

\{prepared\_harness\_summary\}

\#\# Candidate Failure Patterns (arms to score)

Score \textbf{every} pattern below.

\{candidate\_patterns\}

\#\#\# Candidate Table Columns

- `Pattern`: Stable failure-pattern ID used by the `pattern` and `evo-dag` CLIs.
- `Activity`: Rested EMA of the raw observations, seeded at `1.00`. This is a
  mechanical severity reference, not the Agent learning-progress score $\phi$.
- `Observations`: Complete per-iteration observation history. Each entry uses
  `iteration:tagged/evaluated (active=...), tagged=[...], evaluated=[...]`.
  `evaluated` lists the scenarios associated with this pattern that were
  evaluated in that iteration's mini-batch, and `tagged` lists the subset that
  was still tagged with the pattern after patching. `active` is the corresponding
  `tagged/evaluated` fraction. For legacy observations whose scenario IDs were
  not recorded, the counts appear as `?/?` and both lists as `(not recorded)`;
  `(none)` means the arm has not been selected again since the pattern was most
  recently observed in a scenario, so no subsequent activity observation has
  been recorded.
- `\#Scen`: Number of unique scenarios currently tagged with the pattern.
- `LastObs`: Most recent iteration in which the arm was selected and an activity observation was recorded, or `0` if the arm has not been selected again since the pattern was most recently observed in a scenario.
- `Label`: Symptom-level failure-pattern description.
- `Scenarios`: Complete list of scenario IDs currently owned by the arm.

\#\# Workflow
\#\#\# 1. Understand the current harness (history-analysis)

Run the `history-analysis` skill. Inspect the current prepared node with
`evo-dag show node \{prepared\_candidate\_idx\}` and its provisional parent with
`evo-dag show node \{provisional\_parent\_idx\}`. Read the actual prepared-harness change
with `git diff \{provisional\_parent\_commit\} \{prepared\_commit\}`.
You need this context to assess all four scoring axes: *severity* (whether the failure pattern is likely still active under the prepared harness), *fixability* (whether and where a patch can address it), *breadth* (how broadly a successful repair may generalize), and *side-effect risk* (which components or behaviors a patch may affect or regress).

\#\#\# 2. Review prior pulls for each candidate arm

Read the histories of all arms that have previously been selected:

```bash
pattern history
```

With no `--pattern-id`, the command shows every historically pulled arm. If the full output is too large to review efficiently, you may optionally narrow it to a subset in one call with `pattern history --pattern-id <id1> <id2>`, or use `--last-k <K>` to show only the K most recent entries for each selected arm during initial triage. These options are convenience filters, not substitutes for complete review: inspect an arm's full history before rating it whenever older attempts could change your judgment. The output distinguishes patched attempts, all-pass skips, and failed attempts and includes diagnoses, patch intents, dev-set impact, and per-scenario reflections.

\#\#\# 3. Inspect each candidate pattern

For every pattern in the table:
- `pattern show <pattern\_id>` --- label, tagged scenarios, evidence (root causes).
- `evo-dag show scenario <sid>` --- for each of the pattern's scenarios, read its
  failure history and relate it to the current harness change.

\#\#\# 4. Estimate learning progress (progress-scoring)

Follow the `progress-scoring` skill. For each pattern, judge the four
considerations (severity, fixability, breadth, side-effect), then rate
\textbf{all four axes carefully} in [0, 1]. The final score $\phi$ is the mean of those
four axes with side-effect inverted ---
`(severity + fixability + breadth + (1 - side\_effect)) / 4`. Your per-axis
ratings (not a single holistic judgment) determine the result.

**Use the per-arm pull history as primary evidence, and weigh the dev-set, not
just the mini-batch.** A scenario merely passing the mini-batch in a recent pull
does NOT by itself mean the pattern is resolved --- learning progress is measured
on the \textbf{dev/held-out} distribution:
- If a prior patch fixed the scenario but \textbf{dropped dev-set accuracy} (so it was
  rejected and reverted), a better dev-preserving fix is still needed $\rightarrow$ progress
  may remain \textbf{high}.
- If the scenario is genuinely resolved (passing with no dev headroom left), or
  its failure is intrinsically unfixable without regressing dev $\rightarrow$ progress is
  \textbf{low}.
- If recent pulls show the scenarios repeatedly passing, lower the score, **but
  do not force it to 0** --- the pass may be intermittent, and the sampler's
  minimum-probability floor still re-checks low-scored arms periodically to catch
  silent regressions.

\#\#\# 5. Record every score

> \textbf{Reminder}: $\phi$ is computed from your four axes (see step 4). `--side-effect` is a RISK --- higher means worse.

For each candidate pattern:
```bash
pattern rate --pattern-id <id> \textbackslash{}
  --severity <s> --fixability <f> --breadth <b> --side-effect <e> \textbackslash{}
  --rationale "..."
```
\textbf{CRITICAL}: rate every pattern in the table before finishing. An un-rated
pattern is treated as \textbf{low priority (score 0.0)} --- if you skip one it gets
deprioritized rather than reflecting your judgment, so do not leave any
un-rated.

\#\# What Happens Next

The outer loop samples one pattern with probability proportional to `softmax($\phi$ / $\tau$)` and sends its scenarios to the diagnosis-and-patching step.
}
\end{tcolorbox}

\captionof{figure}{
Arm Scoring Prompt: the agent estimates a learning-progress score $\phi_t(a)$ for every instantiated failure-pattern arm by considering severity, fixability, breadth, and side-effect risk. The resulting scores define the stochastic prioritization of known arms.
}
\label{fig:arm-scoring-prompt}

\begin{tcolorbox}[
    enhanced,
    breakable,                       
    colback=white,
    colframe=gray!50,
    boxrule=0.4pt,
    arc=4pt,
    left=8pt, right=8pt, top=4pt, bottom=4pt,
    title=\textbf{Unseen Scenario Exploration Prompt},
    coltitle=white,
    colbacktitle=black,
    fonttitle=\small\bfseries,
    halign title=center,
    attach boxed title to top={yshift=-0.5mm, xshift=-0.4pt},
    boxed title style={
        colback=black,
        colframe=black,
        boxrule=0pt,
        arc=4pt,
        sharp corners=south,
    },
    width=\linewidth,
    pad at break*=2mm,
    overlay middle and last={
        \node[anchor=south east, font=\footnotesize\itshape, gray]
        at ([xshift=-6pt,yshift=2pt]frame.north east)
        {(continued)};
    },
]
{\fontsize{8pt}{9.5pt}\ttfamily%
\setlength{\parindent}{0pt}%
\setlength{\parskip}{2pt}%
\raggedright%
\obeylines%
\#\# Mandatory Skills

You MUST read and follow the SKILL.md for each skill listed below.
Do NOT skip or summarize the skill --- execute the full procedure described in
it. This skill is installed at `.claude/skills/<name>/SKILL.md` in the current
worktree.

| Order | Skill | When | Why |
|-------|-------|------|-----|
| 1 | `history-analysis` | Before the decision (Step 1) | Understand the Prepared harness and evolution history |

\#\# Goal

Decide whether this iteration should exploit a known failure-pattern arm
(PULL) or explore unseen scenarios that may reveal a new failure type (DRAW).
Record exactly one decision with `pattern decide`.

This is an exploration-exploitation trade-off in \textbf{arm space}:

- \textbf{PULL} spends this iteration on an already-discovered pattern. arm-scoring step
  then scores every known pattern so the sampler can choose an arm to repair.
- \textbf{DRAW} probes scenarios that have never been executed. This may reveal a
  failure pattern not yet represented by the known arms; Arm scoring is skipped.

There is \textbf{no fixed formula or threshold} for this decision. Do not decide
mechanically from `|P\_t|`, `|U\_t|`, Activity, or any single statistic. Weigh all
available evidence and make one holistic decision.
Here, `P\_t` is the set of failure patterns discovered by iteration `t`, and `U\_t` is the set of scenarios still unseen at that iteration; `|...|` denotes the size of a set.

This is a \textbf{read-only analysis} session: do NOT modify the codebase and do NOT
rate arms.

\#\# Context

- \textbf{Iteration}: \{iteration\}
- \textbf{Prepared harness}: C\{prepared\_candidate\_idx\}
- \textbf{Prepared worktree}: `\{prepared\_worktree\_path\}`
- \textbf{Prepared commit}: `\{prepared\_commit\}`
- \textbf{Provisional parent}: C\{provisional\_parent\_idx\} (`\{provisional\_parent\_commit\}`)
- \textbf{Selection JSON}: `\{selection\_session\_json\_path\}`
- \textbf{Session root}: `\{session\_root\}`
- \textbf{Discovered failure patterns}: |P\_t| = \{num\_arms\}
- \textbf{Unseen scenarios remaining}: |U\_t| = \{unseen\_pool\_size\}

\#\#\# Prepared Harness Summary

\{prepared\_harness\_summary\}

\#\# Candidate Failure Patterns (known arms)

Use the patterns below to judge whether a worthwhile known target exists.

\{candidate\_patterns\}

\#\#\# Candidate Table Columns

- `Pattern`: Stable failure-pattern ID used by the `pattern` and `evo-dag` CLIs.
- `Activity`: Rested EMA of the raw observations, seeded at `1.00`. This is a
  mechanical severity reference, not the Agent learning-progress score $\phi$.
- `Observations`: Complete per-iteration observation history. Each entry uses
  `iteration:tagged/evaluated (active=...), tagged=[...], evaluated=[...]`.
  `evaluated` lists the scenarios associated with this pattern that were
  evaluated in that iteration's mini-batch, and `tagged` lists the subset that
  was still tagged with the pattern after patching. `active` is the corresponding
  `tagged/evaluated` fraction. For legacy observations whose scenario IDs were
  not recorded, the counts appear as `?/?` and both lists as `(not recorded)`;
  `(none)` means the arm has not been selected again since the pattern was most
  recently observed in a scenario, so no subsequent activity observation has
  been recorded.
- `\#Scen`: Number of unique scenarios currently tagged with the pattern.
- `LastObs`: Most recent iteration in which the arm was selected and an activity observation was recorded, or `0` if the arm has not been selected again since the pattern was most recently observed in a scenario.
- `Label`: Symptom-level failure-pattern description.
- `Scenarios`: Complete list of scenario IDs currently owned by the arm.

\#\# Workflow

\#\#\# 1. Understand the current harness (history-analysis)

Run the `history-analysis` skill. Inspect the current prepared node with
`evo-dag show node \{prepared\_candidate\_idx\}` and its provisional parent with
`evo-dag show node \{provisional\_parent\_idx\}`. Read the actual prepared-harness change
with `git diff \{provisional\_parent\_commit\} \{prepared\_commit\}`.

Use this context to judge whether known-arm evidence still applies to the
prepared harness and whether its weaknesses remain reachable by another patch.

\#\#\# 2. Review prior pulls for each candidate arm

Read the histories of all arms that have previously been selected:

```bash
pattern history
```

With no `--pattern-id`, the command shows every historically pulled arm. If the
full output is too large for initial triage, narrow it with
`pattern history --pattern-id <id1> <id2>` or apply `--last-k <K>`
separately to each selected arm. These are convenience filters, not substitutes
for complete review when older attempts could change the PULL/DRAW decision.

The output distinguishes patched attempts, all-pass skips, and failed attempts
and includes patch approaches, dev-set impact, and per-scenario reflections.

\#\#\# 3. Inspect promising known patterns

Use `pattern show <pattern\_id>` for any pattern that appears impactful. Read its
label, tagged scenarios, root-cause evidence, and observations. You do not need
to assign a learning-progress score in this session; make a coarse judgment of
whether at least one strong known target exists.

\#\#\# 4. Weigh the PULL/DRAW considerations

Use all two lenses below. They guide attention but do not form a formula.

1. \textbf{Is there a clearly worthwhile known pattern to fix?}
   A severe, plausibly fixable, broad pattern argues for PULL. If known patterns
   look near-resolved, unreachable, or exhausted by repeated low-yield attempts,
   diminishing returns argue for DRAW.
2. \textbf{How complete is coverage of the failure surface?}
   Few discovered arms and a large unseen pool suggest that known patterns do
   not yet represent the harness's weaknesses. A small or exhausted unseen pool
   leaves little discovery value and favors PULL.

\#\#\# 5. Make and record one decision

The discovery value of DRAW is a prediction: unseen scenario contents are not
known. Do not over-explore late when a strong fix is available, and do not
over-exploit early when the failure surface is poorly mapped.

Record exactly one action:

```bash
pattern decide --action pull \textbackslash{}
  --rationale "why a known pattern is the best use of this iteration"
\# OR
pattern decide --action draw \textbackslash{}
  --rationale "why discovery is more valuable than fixing known patterns now"
```

Do NOT call `pattern rate` in this session.

\#\# What Happens Next

- After PULL, arm scoring runs on the same prepared harness and scores every arm.
- After DRAW, arm scoring is skipped and the outer loop samples unseen scenarios.
}
\end{tcolorbox}

\captionof{figure}{
Unseen Scenario Exploration Prompt: the agent decides whether to continue optimizing a known failure-pattern arm or explore scenarios that may reveal new failure patterns by weighing the value of unresolved known targets against the remaining value of further exploration.
}
\label{fig:unseen-scenario-exploration-prompt}

\section{Limitations and Responsible Deployment}
\label{sec:limitations}

\proposed{} is a research exploration of automated curriculum learning for
offline harness optimization and is not intended or evaluated as a
production-ready agent system.
Our experiments are conducted entirely in benchmark environments using public and
synthetic task data; no real user data are used in the study.
Accordingly, the empirical results characterize optimization effectiveness
under these controlled settings rather than production safety, security,
privacy, or governance readiness.

Improved benchmark performance should therefore not be interpreted as evidence
that an optimized harness is safe or compliant for unrestricted deployment.
This distinction is important for harness optimization because
updates may modify prompts, tool interfaces, middleware, and runtime control
logic, thereby changing not only task behavior but also how an agent accesses
resources and executes actions.
Moreover, \proposed{} prioritizes optimization opportunities according to
observed task failures and the objectives provided by the underlying harness
optimizer; it does not by itself determine whether the resulting behavior
satisfies independent safety, security, privacy, or policy requirements.

\paragraph{Safety and responsible AI.}
Because \proposed{} is evaluated as a research optimization method rather than
a production agent system, our experiments do not include dedicated
evaluations of agent misuse, cyber-abuse behavior, or other unsafe deployment
scenarios.
We also do not introduce an independent safety filter or a dedicated mechanism
for detecting reward hacking or specification gaming, where an optimizer
improves the measured objective without satisfying its intended semantics.
Accordingly, the task-level objectives and failure signals used by
\proposed{} should not be interpreted as a complete specification of acceptable
agent behavior.
A production deployment would need to augment task-level optimization with
independent safety criteria, adversarial misuse and policy-violation testing,
and safety regression checks throughout the optimization and release process.
Safety-relevant signals should be treated as independent constraints or review
criteria rather than inferred solely from improvements in task performance.

\paragraph{Security.}
Our study does not evaluate production security risks such as unsafe command
execution, privilege misuse, unauthorized access to sensitive files, or
escalation through tools.
These risks are particularly relevant to harness optimization because an
automatically generated update may modify prompts, tool interfaces, middleware,
command construction, tool-selection logic, or runtime control flow even when
the underlying model weights remain unchanged.
In a production setting, such changes should therefore be evaluated within
explicit security boundaries, including least-privilege tool access,
sandboxing, sensitive-resource isolation, and adversarial testing of tool and
command execution.
Updates that modify permissions, exposed capabilities, tool interfaces, or
other security-critical behavior should require explicit review and should not
be promoted to production solely on the basis of task-performance improvements.
Deployment mechanisms should additionally support rollback when security
regressions are detected.

\paragraph{Privacy.}
Because our experiments use public/synthetic  benchmarks and do not involve real
user data, they do not exercise the privacy risks that can arise when
optimization is applied to production execution traces.
In real deployments, however, traces used for failure diagnosis and harness
optimization may contain personally identifiable information (PII),
credentials, proprietary content, or other sensitive data.
Such deployments should therefore minimize the data collected for optimization
and apply PII and secret detection, sensitive-content filtering, access
controls, and anonymization or redaction where appropriate before traces are
persisted or consumed by the optimization pipeline.
They should also define an explicit, purpose-limited retention period
consistent with the applicable data classification and privacy requirements,
rather than retaining optimization traces indefinitely.
These production privacy controls are outside the scope of the present
benchmark study.

\paragraph{Governance and deployment controls.}
Automatic acceptance or promotion of an update by \proposed{} should not
itself constitute approval for production deployment.
Organizations applying automated harness optimization in production should
conduct the applicable AI impact and risk assessments and maintain governance
processes that are independent of the optimizer's task-level objective.
Depending on the deployment context, these processes may include
human-in-the-loop approval for high-impact or security-sensitive changes,
auditable records of generated, evaluated, accepted, and rejected updates,
staged evaluation before release, explicit safety-review and risk-acceptance
checkpoints, and defined rollback procedures when regressions are detected.
The required degree of oversight should depend on the capabilities exposed to
the agent, the sensitivity of the affected resources, and the potential impact
of its actions.
More generally, optimizing task performance does not remove the need for
deployment-specific safety, security, privacy, and compliance review.

\end{document}